\documentclass[11pt]{article}

\usepackage[english]{babel}
\usepackage[T1]{fontenc}

\usepackage[a4paper,top=2.75cm,bottom=2.75cm,left=2.75cm,right=2.75cm,marginparwidth=1.75cm]{geometry}
\usepackage{amsmath}
\usepackage{amsthm}
\usepackage{amssymb}
\usepackage{bbm}
\usepackage{IEEEtrantools}
\usepackage[]{mathrsfs}
\usepackage{siunitx}
\usepackage{graphicx}
\usepackage[colorinlistoftodos]{todonotes}
\usepackage[colorlinks=true, allcolors=blue]{hyperref}
\usepackage{bm}

\usepackage{enumitem}

\usepackage{booktabs,multirow,array}

\theoremstyle{definition}

\usepackage{graphicx, setspace, latexsym,amsmath,amssymb,color}
\usepackage{amssymb} 
\usepackage{amsmath} 
\usepackage{amsthm}
\usepackage{bm}

\usepackage{quoting}
\usepackage{dsfont}
\usepackage{longtable}
\usepackage{lineno}
\usepackage{parskip}
\usepackage[normalem]{ulem}

\usepackage{xcolor}
\usepackage{tikz}
\usepackage{wrapfig}
\usepackage{tocbibind}

\usepackage{amsmath}
\usepackage{amsthm}
\usepackage{amssymb}
\usepackage{bbm}
\usepackage{IEEEtrantools}
\usepackage[]{mathrsfs}
\usepackage{siunitx}
\usepackage{graphicx}
\usepackage{wrapfig}
\usepackage{subfigure}
\usepackage{algorithm}
\usepackage{algorithmic}

\usepackage{enumitem}
\usepackage{tcolorbox}
\usepackage{booktabs,multirow,array}
\usepackage{optidef}
\newtheorem{lemma}{Lemma}

\newtheorem{prop}{Proposition}
\usepackage{tabto}

\newtheorem{theorem}{Theorem}

\newtheorem{remark}{Remark}

\def\bx{{\bm x}}
\def\by{{\bf y}}

\def\bc{{\bf c}}

\newcommand{\Lb}{{\bf L}}

\newcommand{\Ab}{{\bf A}}
\newcommand{\Wb}{{\bf W}}

\newcommand{\Gb}{{\bf G}}
\newcommand{\Kb}{{\bf K}}
\newcommand{\Mb}{{\bf M}}

\newcommand{\Ub}{{\bf U}}

\newcommand{\xb}{{\bm x}}

\title{
\textbf{Metric Learning in an RKHS}}
\author{Gokcan Tatli$^\dagger$, Yi Chen$^\dagger$, Blake Mason$^\ddagger$,  Robert Nowak$^\dagger$, Ramya Korlakai Vinayak$^\dagger$\\ \\ $^\dagger$Department of Electrical and Computer Engineering, University of Wisconsin-Madison\\ $^\ddagger$Amazon.com, USA\\
\tt{\{gtatli@, yi.chen@, rdnowak@, ramya@ece.\}wisc.edu, bjmason@amazon.com}}
\begin{document}
\date{}
\maketitle
\begin{abstract}
  Metric learning from a set of triplet comparisons in the form of ``\textit{Do you think item h is more similar to item i or item j?}'', indicating similarity and differences between items, plays a key role in various applications including image retrieval, recommendation systems, and cognitive psychology. The goal is to learn a metric in the RKHS that reflects the comparisons. Nonlinear metric learning using kernel methods and neural networks have shown great empirical promise. While previous works have addressed certain aspects of this problem, there is little or no theoretical understanding of such methods. The exception is the special (linear) case in which the RKHS is the standard Euclidean space $\mathbb{R}^d$; there is a comprehensive theory for metric learning in $\mathbb{R}^d$. This paper develops a general RKHS framework for metric learning and provides novel generalization guarantees and sample complexity bounds. We validate our findings through a set of simulations and experiments on real datasets. Our code is publicly available at \texttt{\url{https://github.com/RamyaLab/metric-learning-RKHS}}.
\end{abstract}

\addtocontents{toc}{\protect\setcounter{tocdepth}{0}}
\section{Introduction}
\begin{figure*}[t]
    \centering
    \includegraphics[width=1\linewidth]{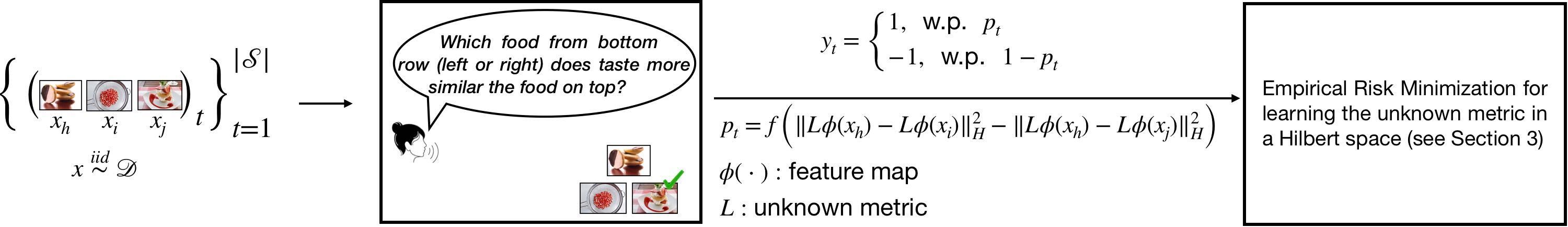}
    \caption{Metric Learning from triplet comparisons (example triplets from Food-100 dataset \cite{wilber2014cost}). $\mathcal{S}$ is the set of triplets and $y_t$ is the label collected from human for each triplet $t$.}
    \label{fig:triplets}
\end{figure*}

Understanding how humans perceive objects is essential in many areas from machine learning \cite{hu2015deep, hsieh2017collaborative} to psychology \cite{cao2013similarity, roads2019obtaining} and policy learning \cite{liu2021deep}. Learning representations over objects that reflects similarities and dissimilarities on human perception is key to this understanding. Metric learning focuses on the problem of learning a distance function that represents similarities and dissimilarities among objects. This is particularly useful in computer vision applications such as image retrieval \cite{hoi2010semi, yao2020adaptive} and face recognition \cite{guillaumin2009you, cao2013similarity}, and recommendation systems \cite{zhang2019next, wu2020effective}, where the notion of similarity plays a central role on the performance. Comparative judgments over objects has been widely used as a powerful tool in these applications and many others to understand similarities and dissimilarities. In this paper, we provide a theoretical foundation to the task of metric learning in RKHS from triplet comparisons in the form of \textit{“is item h more similar to item i or to item j?”} (see Figure \ref{fig:triplets} for an example triplet comparison query for Food-100 dataset \cite{wilber2014cost}). The goal is to learn a metric that predicts triplet comparisons as well as possible by learning a distance function. Let $\bx\in\mathbb{R}^d$ be the representation of objects. We are given a random set of triplet comparisons which compare relative distances between a head item $\bx_h$ to two alternates $\bx_i, \bx_j$ in the form of 
\begin{eqnarray*}
    \text{sign}(\text{dist}^2(\bm{x}_h,\bm{x}_i)-\text{dist}^2(\bm{x}_h,\bm{x}_j)).
\end{eqnarray*}
As an example, items may be images of products sold in an online marketplace and features $\bx_i$ could either be constructed from metadata about each product or extracted automatically from the image via a pre-trained neural network. As human judgments are complex and involve higher order interactions of features, we seek a sufficiently expressive family of distance metrics to model these judgments. Hence, we consider learning a nonlinear metric represented with a kernelized setting. 

In the special case of a linear kernel, this problem corresponds to learning the Mahalanobis metric represented by a positive semidefinite matrix $\Mb$. A triplet comparison in the linear setting then takes the following form,
\begin{eqnarray*}
\text{sign}\left( \|\bm{x}_h-\bm{x}_i\|^2_{\textbf{M}}-\|\bm{x}_h-\bm{x}_j\|^2_{\textbf{M}} \right).
\end{eqnarray*}
Since $\Mb$ is positive semidefinite, we can express it as $\Mb=\Lb^T\Lb$ using the Cholesky decomposition. Thus, learning the positive semidefinite matrix $\Mb$ is equivalent to learning a linear transformation $\Lb$ such that the distances are interpreted as Euclidean distances between points transformed by the matrix $\Lb$. This linear setting has been well studied \cite{verma2015sample, mason2017learning, ye2019fast}. While nonlinear metric learning using kernel methods and neural networks has been widely applied in practice, its theoretical analysis remains limited. Our work extends the theoretical understanding from linear metric learning framework to the kernelized scenario, where the focus is on learning a linear metric on a reproducing kernel Hilbert space (RKHS). 

We assume that we have access to a feature map $\phi$ that maps from $\mathbb{R}^d$ to a real reproducing kernel Hilbert space (RKHS) $\mathcal{H}$ such that $\langle \phi(\bx_i), \phi(\bx_j)\rangle=k(\bx_i, \bx_j)$ and $\|\phi(\bx)\|_
\mathcal{H}=\sqrt{k(\bx, \bx)}$ for a known kernel function $k: \mathbb{R}^d\times \mathbb{R}^d \rightarrow \mathbb{R}^1$. Therefore, $k(\cdot, \cdot)$ satisfies the reproducing property that $\langle f, k(\cdot, \bx)\rangle=f(\xb)$ for any $f\in \mathcal{H}$ and $\bx \in \mathbb{R}^d$. Then for any bounded linear operator $L: \mathcal{H}\rightarrow \mathcal{H}$, we define an associated nonlinear Mahalanobis metric, $d_L$, as 
\begin{eqnarray*}
    d_L^2 (\xb_i, \xb_j) &=&\|L\phi(\xb_i)-L\phi(\xb_j)\|_\mathcal{H}^2 \\
    &=&{\langle L\phi(\xb_i)-L\phi(\xb_j), L\phi(\xb_i)-L\phi(\xb_j) \rangle_\mathcal{H}}.
\end{eqnarray*}
For simplicity, we use $\phi_i$ for $\phi(\bx_i)$ for the rest of the paper. With the kernelized metric setting, we can write the response to a triplet query as
\begin{eqnarray*}
\text{sign}\left( \|L\phi_h-L\phi_i\|_\mathcal{H}^2 -\|L\phi_h-L\phi_j\|_\mathcal{H}^2 \right).
\end{eqnarray*}
\textbf{Our Contributions:} This paper advances the theoretical understanding of the empirically powerful tasks of nonlinear metric learning through the following contributions:
\begin{itemize}
    \item We establish the first generalization error and sample complexity guarantees for kernelized metric learning from triplet comparisons. 
    \item We provide insights into how regularization affects the sample complexity and generalization bounds for kernelized metric learning from triplet comparisons. 
    \item As a byproduct, our analysis extends the results of the linear metric learning setting \cite{mason2017learning}, overcoming its limited applicability, which required the number of items $n$ to be larger than the dimensionality $d$. 
\end{itemize}

\subsection{Related Works}
Metric learning (also known as distance learning) has gained significant interest due to its power of effectively learning similarities and dissimilarities among objects. Here, we summarize most relevant contributions from the rich literature on metric learning. There exist comprehensive summaries of the literature on classical techniques \cite{kulis2013metric,bellet2015metric}. In this paper, our focus is on a specific type of query known as triplet comparisons. There are methods and efficient algorithms for a variety of feedback such as class labels \cite{weinberger2009distance,davis2007information}, triplet comparisons \cite{schultz2003learning,mason2017learning}, perceptual adjustment queries \cite{xu2024perceptual} and nearest neighbor queries \cite{nadagouda2023active}. A recent study \cite{tatli2024metric} uses triplet comparison queries to learn heterogeneous metrics by first clustering users, enabling the discovery of latent subgroups within the population before proceeding to metric learning within each subgroup. 

Verma and Branson \cite{verma2015sample} provide sample complexity of Mahalanobis
distance learning from class labels, which is also known as linear metric learning, where the metric is parametrized by a positive semidefinite matrix. Later works \cite{mason2017learning, ye2019fast} present tight generalization error bounds for Mahalanobis distance metric learning from triplet comparisons. Recently, there has been increased interest in nonlinear metric learning to better fit complex, real-world data sources. Kernelized approaches to the metric learning, similar to the setting considered in this work, are proposed in several studies (see references for a subset of them \cite{martinel2015kernelized, liu2021fast, wang2011metric, chatpatanasiri2010new,kleindessner2017kernel}). More generally, the nonlinear variant has received attention through the study of deep Siamese networks \cite{guo2017learning}.

Recent interest in using deep learning to extract useful representations from  data is followed by triplet network models and its variations \cite{hoffer2015deep}. Kaya and Bilge \cite{kaya2019deep} provide a comprehensive survey on deep metric learning. Despite the empirical success and popularity of deep metric learning techniques on metric learning, theoretical advancements in this area remain sparse. A recent work \cite{zhou2024generalization} provides a generalization analysis with deep ReLU networks for metric learning using the hinge loss. Other studies provide generalization guarantees for deep metric learning using neural tangent kernel \cite{liu2021fast} and using Rademacher complexity based analysis \cite{huai2019deep}.

Another line of work focuses on metric learning from pairwise comparisons. Pairwise comparisons can be viewed as a variation of triplet comparisons when it is assumed that there is a reference point $u$ (responder) substituting for leading item $\bx_h$. How to infer preferences from pairwise comparisons is a well studied problem in a diverse set of areas including machine learning, social choice theory, psychology, social sciences and political science (see the reference \cite{furnkranz2010preference} for a comprehensive summary). Xu and Davenport \cite{xu2020simultaneous} propose a passive algorithm to learn a linear metric and preferences. This can be seen as simultaneously performing metric and preference learning. Later, another work \cite{canal2022one} extends the results to learning multiple preference points with a shared metric and provides theoretical guarantees for this task. A recent study \cite{wang2024metric} analyzes the linear metric learning problem with limited pairwise comparisons per user. Another recent study \cite{chen2025pal} proposes leveraging preference structure to reduce the sample complexity of pairwise comparisons for simultaneous metric and preference learning with applications to alignment.

Another line of work considers a more discrete task of clustering items based on answers to comparison queries~\cite{gomes2011crowdclustering,korlakai2014graph, korlakai2016crowdsourced, mazumdar2017clustering, mazumdar2017semisupervised, vinayak2017tensor, mazumdar2017semisupervised, ibrahim2019crowdsourcing,ibrahim2021mixed, chen2023crowdsourced, chen2025query}. The goal in these works is to assign items to clusters rather than learning a metric.

\section{Problem Setting}
Let objects be represented by the points $\bm{x}_1, \bm{x}_2, \ldots$, where each $\bm{x}_i$ is drawn from the distribution $\mathcal{D}'$. In the noiseless setting, we are given a set of triplet comparisons in the form of
\begin{eqnarray*}
    \text{sign}(\text{dist}^2(\bm{x}_h,\bm{x}_i)-\text{dist}^2(\bm{x}_h,\bm{x}_j)).
\end{eqnarray*}
We are interested in providing a theoretical understanding on the problem of learning kernelized Mahalanobis metric from triplet comparison queries. Our work extends the learning theoretic results of a previous work on linear metric learning \cite{mason2017learning} to more general nonlinear metrics.

Let $\mathcal{S}$ denote the set of triplets generated from random triples $t=\{\bx_{h}, \bx_{i}, \bx_{j}\}$, where each triple is independent and randomly chosen from the distribution $\mathcal{D}$, i.e., given that $\bx_i\sim \mathcal{D}'$, each triple $t_{\{h,i,j\}}\in \mathcal{S}$ is randomly sampled from the stacked distribution $\mathcal{D}$. Therefore, the total number of objects is $3|\mathcal{S}|$ for $|\mathcal{S}|$ triplets in the general case. For each random triplet $t_{\{h,i,j\}}$, we observe a possibly noisy answer $y_t\in\{\pm 1\}$, which is an indication of $\text{sign}\left(\|L\phi_h-L\phi_i\|_\mathcal{H}^2-\|L\phi_h-L\phi_j\|_\mathcal{H}^2\right)$. Specifically, we assume that there exists an unknown kernelized metric that is consistent with the data and classifies any triplet $t$ correctly with a probability greater than $1/2$ where this probability is taken with respect to any randomness in $y_t$ and may depend on the specific triplet $t$. This is a common practical assumption when working with human judgment that some queries are inherently more noisy than others \cite{coombs1964theory,rau2016model}. We further assume that the $y_t'$s are statistically independent. Our goal is to learn a metric parameterized by a linear map $L$ that predicts triplets well on average. Namely, we seek an $L$ that minimizes the misclassification probability:
\begin{eqnarray}
    \text{Pr}\left(y_t\neq \text{sign}\left(\|L\phi_h-L\phi_i\|_\mathcal{H}^2-\|L\phi_h-L\phi_j\|_\mathcal{H}^2\right)\right).\label{0-1 loss}
\end{eqnarray}
Note that (\ref{0-1 loss}) is equal to the expected $0/1$ loss. In practice, minimizing $0/1$-loss is intractable and the above objective is relaxed to minimizing the true risk, which is defined below:
  \begin{eqnarray}\label{true_risk}
    {R}(L):= 
    \mathbb{E}_{t\sim \mathcal{D}, y_t \in \{\pm 1\}}[l(y_t(\|L\phi_h-L\phi_i\|_\mathcal{H}^2-\|L\phi_h-L\phi_j\|_\mathcal{H}^2))], 
\end{eqnarray}
for an arbitrary convex and $\alpha$-Lipschitz loss $\ell : \mathbb{R}\rightarrow \mathbb{R}_{\geq 0}$,
where the expectation is over random triplet coming from a distribution $\mathcal{D}$ and binary random label $y_t$ conditioned on $t$, where $t=\{\bx_h, \bx_i, \bx_j\}$ and $\{\bx_h, \bx_i, \bx_j\}\sim \mathcal{D}$. If $\ell$ is chosen to upper bound the $0/1$-loss (e.g., the hinge loss $\ell(z) = \max(1-z, 0)$ or the logistic loss $\ell(z) = \log(1 + \exp^{-z})$, then $R(L)$ upper bounds the misclassification probability.

Unfortunately, we cannot minimize $R(L)$ directly as the joint distribution of $(t, y_t)$ is unknown. Instead, given a set of triplets $\mathcal{S}$ and their labels $y_t$, we wish to learn a kernelized metric parameterized by a bounded linear map $L : \mathcal{H} \rightarrow \mathcal{H}$ that predicts triplets as well as possible on the observed data. 
\begin{eqnarray}\label{empirical risk first}
  \widehat{R}_\mathcal{S}(L):= 
   \frac{1}{|\mathcal{S}|}\sum_{(t,y_t)\in \mathcal{S}}l(y_t(\|L\phi_h-L\phi_i\|_\mathcal{H}^2-\|L\phi_h-L\phi_j\|_\mathcal{H}^2)).  
\end{eqnarray}
  We refer to $\widehat{R}_\mathcal{S}(L)$ as the empirical risk as it is an unbiased estimator of the true risk $R(L)$. For any given $\ell$, we wish to answer three questions:
  \begin{enumerate}    
      \item Regularizing a norm on $L$ controls the flexibility of the metric and hence the model’s predictions. What is the appropriate way to regularize to balance the bias-variance tradeoff of metric learning?
      \item What can we guarantee about the generalization performance of the solution to (\ref{empirical risk first}) and how does this depend on the norm we choose to regularize on $L$?
        \item As written, (\ref{empirical risk first}) is a potentially infinite dimensional, nonconvex optimization problem. How can it be made computationally tractable?
  \end{enumerate}
We refer to Section \ref{sec:Theoretical Guarantees} for the first and second questions, and Section \ref{sec:practical} for the last question.       
  
\section{Kernelized Metric Learning}\label{KernelizedML}
Traditional Mahalanobis distance metric learning is equivalent to learning a linear mapping of the data such that Euclidean distance in the mapped space agrees with a set of labels, such as class labels or triplet comparisons. Often, we are interested in a richer set of mappings than linear ones. Indeed, this is the idea that underlies deep learning and kernel learning. In this section, we provide the first theoretical study of nonlinear metric learning in an RKHS from triplet data, extending the linear results of a previous work \cite{mason2017learning}. 

\subsection{Theoretical Guarantees for Kernelized Metric Learning}\label{sec:Theoretical Guarantees}
Frequently in optimization and learning theory, we wish to characterize \textit{model classes of functions---} classes of metrics on $\mathcal{H}$ in this case. This is important to define optimal performance within a class for theoretical results and has tight connections to regularization in optimization which is used to prevent overfitting the data and ensure good generalization performance. We define model classes of kernelized Mahalanobis metrics by bounding the Schatten $p$-norm of their map $L$ (e.g., all kernelized metrics with a map $L$ such that $\|L\|_{S_p}\leq \lambda$). For a compact, bounded linear operator ${T}$, its Schatten $p$-norm is defined to be $\|T\|_{S_p}:=\left(\sum_{i\geq 1}s_i\left(T\right)^p \right)^{1/p}$ where $s_i(T)$ is the $i^{th}$ singular value of $T$ and may be equivalently written as $\sqrt{\lambda_i(T^\dagger T)}$ where $\dagger$ denotes the conjugate transpose and $\lambda_i(T^\dagger T)$ is the
$i^{th}$ eigenvalue of the Hermitian operator. We focus on two particular Schatten norms. First we consider the Schatten 2-norm which is a Hilbert-Schmidt norm.  Specifically, we restrict solutions $L$ to (\ref{empirical risk first}) to additionally satisfy $\|L^\dagger L\|_{S_2}\leq \lambda_F$ for a given $\lambda_F>0$. Furthermore, we consider the Schatten 1-norm, also known as the trace or nuclear norm. In this setting, we assume that $\|L^\dagger L\|_{S_1}\leq\lambda_*$ and again restrict solutions to satisfy this constraint. 

We define the optimal (possibly) infinite dimensional operator $L^*$ as the minimizer of following optimization:
\begin{equation}
\begin{aligned}
\min_{L} \quad & R(L)\\
\textrm{s.t.} \quad & \|L^\dagger L\|_{S_2}\leq \lambda_F.    \\
\end{aligned} \tag{P1}\label{opt-P1}
\end{equation}
Similarly we define $\widehat{L}$ as the solution to the optimization problem (\ref{opt-P2}) given below, i.e., the empirical risk minimizer:
\begin{equation}
\begin{aligned}
\min_{L} \quad & \widehat{R}_\mathcal{S}(L)\\
\textrm{s.t.} \quad &  \|L^\dagger L\|_{S_2}\leq \lambda_F    \\
\end{aligned}\tag{P2}\label{opt-P2}
\end{equation}
Suppose that $\mathcal{S}_\mathcal{X}\subset\mathcal{H}$ represents the subspace spanned by the set $\{\phi(\xb_1), \phi(\xb_2), \ldots \phi(\xb_{n})\}$ corresponding to the features of random observations. Furthermore, let the potentially infinite dimensional linear operator $\widehat{L}_0$ denote the solution to (\ref{opt-P3}), obtained from random observations and the associated kernel features, where the norm constraint is imposed solely on the component of $L$ whose domain lies within the span of features, i.e., denoted by $\mathcal{S}_\mathcal{X}$:
\begin{equation}
\begin{aligned}
\min_{L} \quad & \widehat{R}_\mathcal{S}(L)\\
\textrm{s.t.} \quad & \|\mathcal{P}_{\mathcal{S}_\mathcal{X}}^\dagger L^\dagger L\mathcal{P}_{\mathcal{S}_\mathcal{X}}\|_{S_2}\leq \lambda_F,    \\
\end{aligned} \tag{P3}\label{opt-P3}
\end{equation}
where $\mathcal{P}_{\mathcal{S}_\mathcal{X}}$ denotes the projection onto $\mathcal{S}_\mathcal{X}$.
\begin{remark}\label{remark:emp risk in the span}
    Assume $\widehat{L}_0$ denotes the solution of (\ref{opt-P3}) whose domain is restricted to the span of features, i.e., $\mathcal{S}_\mathcal{X}$. This is a reasonable assumption, because $L_0\mathcal{P}_{\mathcal{S}_\mathcal{X}}$ also optimally solves (\ref{opt-P3}) for any solution $L_0$. Therefore, optimizing (\ref{opt-P3}) can be interpreted as seeking such an $\widehat{L}_0$. 
\end{remark}
\begin{lemma} \label{lem:norm inequality}
    Recall that for a compact, bounded linear operator ${T}$, its Schatten $p$-norm is denoted as $\|T\|_{S_p}$. We have, for $p\geq 1$,
\begin{eqnarray*} \|\mathcal{P}^\dagger_{\mathcal{S}_\mathcal{X}}L^\dagger L\mathcal{P}_{\mathcal{S}_\mathcal{X}}\|_{S_p} \leq \|L^\dagger L\|_{S_p}.
\end{eqnarray*}
Note that Schatten $2$-norm is the Hilbert-Schmidt norm.
\end{lemma}
Lemma \ref{lem:norm inequality} allows us to establish a relation between solutions of (\ref{opt-P2}) and (\ref{opt-P3}), as formalized in Proposition \ref{prop:opt2 and opt3}. Note that optimization settings (\ref{opt-P2}) and (\ref{opt-P3}) have the same objective function. The distinction lies in the the norm constraint $\|\cdot\|_{S_2}$ imposed on $L$. In (\ref{opt-P3}), the constraint applies only to the component of $L$ whose domain is restricted to the span of features, denoted by $\mathcal{S}_\mathcal{X}$. Consequently, solving (\ref{opt-P3}) for an operator $\widehat{L}_0$, as in Remark \ref{remark:emp risk in the span}, corresponds to minimizing the empirical risk in (\ref{opt-P2}) under the additional constraint that the search is restricted to the span $\mathcal{S}_\mathcal{X}$.  

\begin{prop} \label{prop:opt2 and opt3}
     We observe that $\widehat{L}_0$ is in the solution set of (\ref{opt-P2}). More precisely, any ${L}$ within the feasible set of (\ref{opt-P2}) is an optimal solution, provided that ${L}\mathcal{P}_{\mathcal{S}_\mathcal{X}}=\widehat{L}_0$. As a result, (\ref{opt-P2}) and (\ref{opt-P3}) have the same optimal value, i.e., 
    \begin{eqnarray*}
        \widehat{{R}}_\mathcal{S}(\widehat{L})= \widehat{{R}}_\mathcal{S}(\widehat{L}_0).
    \end{eqnarray*}
    Therefore, optimizing the empirical risk in (\ref{opt-P2}) with a search restricted to $\mathcal{S}_\mathcal{X}$ suffices to assign optimal value for (\ref{opt-P2}). 
\end{prop}
Recall that we wish to
learn a kernelized metric parametrized by a bounded
linear map $L: \mathcal{H}\rightarrow \mathcal{H}$ that predicts triplets effectively based on random observations. We establish a bound on the generalization error of $\widehat{L}_0$, which is a solution to the empirical risk minimization. Note that $\widehat{L}_0$ solves both (\ref{opt-P3}) and (\ref{opt-P2}). We compare it to the operator $L^*$, which minimizes the true risk.

The following theorem demonstrates that, with a sufficiently large set of triplets $\mathcal{S}$, the performance of $\widehat{L}_0$ is nearly as good as that of $L^*$.
\begin{theorem}\label{thm:generalization_error_withbounded_Fro_norm}
    Fix $\delta, \lambda_F>0$ and let $\ell$ be $\alpha$-Lipschitz. Assume $\|\phi(\bx)\|_\mathcal{H}\leq B$ for any $\bx$. Then, with probability at least $1-\delta$,
\begin{eqnarray*}
    {R}(\widehat{L}_0)-R(L^*)\leq 4\alpha B^2\lambda_F\sqrt{\frac{6}{|\mathcal{S}|}}+12\alpha B^2\lambda_F\sqrt{\frac{2\ln{2/\delta}}{|\mathcal{S}|}}
\end{eqnarray*}
\end{theorem}
For any loss $\ell(\cdot)$ which upper bounds the $0/1-$loss, such as the logistic or hinge losses, the left hand side is an upper bound on the expected prediction accuracy for predicting triplets. Hence, the above result also provides a generalization error guarantee for prediction accuracy. 

To further interpret the result of Theorem \ref{thm:generalization_error_withbounded_Fro_norm}, consider the case of a linear kernel where the points used to generate triplets live in the unit ball in 
$\mathbb{R}^d$. In this case,
one can directly learn $L^TL=\Mb \in \mathbb{R}^{d\times d}$. 
Setting $\lambda_F=O(d)$,
which is sufficient to ensure that the average entry of $\Mb$ is dimensionless, Theorem \ref{thm:generalization_error_withbounded_Fro_norm} shows that sampling $O(d^2\log(1/\delta))$ triplets is sufficient to ensure good generalization. As the number of degrees of freedom for a $d\times d$ matrix is $d^2$, this matches
intuition that the sample complexity should scale with degrees of freedom. In general, $\|L^\dagger L\|_{S_2}$
behaves like a notion of the effective dimensionality $d_\textrm{eff}$ of $L$ \cite{zhang2005learning}. Indeed, if fewer eigenvalues of $L^\dagger L$ are large, then $\lambda_F$ is smaller and the space is nearly low dimensional. Hence, we may interpret Theorem \ref{thm:generalization_error_withbounded_Fro_norm} as suggesting a sample complexity of $O(d_\textrm{eff}^2\log(1/\delta))$. 

Next, we bound the excess risk under the constraint $\|L^\dagger L\|_{S_1}\leq \lambda_*$. Specifically, consider the optimization problems (\ref{opt-P1}), (\ref{opt-P2}) and (\ref{opt-P3}), now with Schatten 1-norm constraints of the form $\|\cdot\|_{S_1}\leq \lambda_*$. Let $L^*_n $, \( \widehat{L}_n \) and \( \widehat{L}_{n_0} \) denote the solutions to the modified versions of problems (\ref{opt-P1}), (\ref{opt-P2}) and (\ref{opt-P3}), respectively, where the Schatten 1-norm constraints replace the Schatten 2-norm constraints. 

The following theorem establishes a bound on the generalization error of $\widehat{L}_{n_0}$ by comparing it to the true risk minimizer $L^*_n$.
\begin{theorem}\label{thm:generalization_error_withbounded_Nuclear_norm}
 Fix $\delta, \lambda_*>0$ and let $\ell$ be $\alpha$-Lipschitz. Assume $\|\phi(\bx)\|_\mathcal{H}\leq B$ for any $\bx$. Then, with probability at least $1-\delta$,
\begin{eqnarray*}
    {R}(\widehat{L}_{n_0})-R(L^*_n)\leq 4\alpha\lambda_*\left(B^2\sqrt{12\frac{\log 3|\mathcal{S}|}{|\mathcal{S}|} } +\frac{2\log 3|\mathcal{S}|}{|\mathcal{S}|}\right)
    +12\alpha B^2\lambda_*\sqrt{\frac{2\ln{2/\delta}}{|\mathcal{S}|}},
\end{eqnarray*}
\end{theorem}
where $\mathcal{S}$ is the set of triplets chosen and $|\mathcal{S}|$ represents the size of this set. Note that restricting the Schatten 1-norm encourages solution $L$ (and correspondingly operationalized version $\Mb$ (see Section \ref{sec: learning M})) to have low rank. This corresponds
to learning a low-dimensional metric over data. This is reasonable in settings where though the ambient dimension of data is large, one expects that the triplet comparisons are well explained by a projection of the data points onto a low dimensional space $\mathcal{S}^o$. As an example, consider $\phi$ corresponding to a polynomial kernel of degree 2: $\phi(\bx) = [\xb_1^2, \xb_1 \cdot \xb_2,\ldots,\bx_2^2, \bx_2 \cdot \bx_3, \ldots, \bx_d^2]^T$ for $\xb=[\bx_1,\ldots , \bx_d]^T$. Suppose the data is generated according to a true map $L^*$ which is a projection onto $\mathcal{S}^o$, the span of a sparse subset of $k\ll d^2$ monomials. Then, taking $\lambda_*=\|L^\dagger L\|_{S_1}=k$, Theorem \ref{thm:generalization_error_withbounded_Nuclear_norm} guarantees that sampling $O(k^2\log(k/\delta))$ triplets is sufficient. By contrast, if $L$ was the identity map on degree 2 polynomials, the same result would suggest a sample complexity of $O(d^4\log(d/\delta))$ which is much larger. Hence, this result is especially powerful for low or approximately low dimensional metrics.

\section{Practical Implementation} \label{sec:practical}
In Section \ref{KernelizedML}, we show that solving (\ref{opt-P2}) with a
search restricted to $\mathcal{S}_\mathcal{X}$, i.e., solving for $\widehat{L}_0$ in (\ref{opt-P3}), presents a solution for both (\ref{opt-P2}) and (\ref{opt-P3}). We bound the generalization error based on $\widehat{L}_0$ (see Theorems \ref{thm:generalization_error_withbounded_Fro_norm} and \ref{thm:generalization_error_withbounded_Nuclear_norm}). Our goal in this part is to solve (\ref{opt-P3}) to learn $\widehat{L}_0$, which is a nonlinear Mahalanobis metric. Note that in addition to being possibly infinite dimensional, the optimization (\ref{opt-P3}) is also nonconvex.  



In this section, we carefully demonstrate how to learn $\widehat{L}_0$ from a random set of independent triplets $\mathcal{S}$ with associated labels $y_t$ via convex optimization. We show that solving ($\ref{opt-P3}$) is equivalent to solving a finite dimensional convex optimization problem. We use a representer theorem (see Proposition \ref{prop:representer theorem}) to reduce finding $\widehat{L}_0$ to an optimization over finite dimensional vectors. We use the idea of Kernelized Principle Component Analysis (KPCA) to compute all distances using KPCA vectors $\varphi_1, \varphi_2, \ldots, \varphi_n\in \mathbb{R}^{n}$ and reduce the problem to learning an $n-$dimensional metric parameterized by a semidefinite matrix denoted by $\Mb$:
 \begin{equation}
\begin{aligned}
\widehat{\overline{R}}_\mathcal{S}(\Mb) := \frac{1}{|\mathcal{S}|}\sum_{(t,y_t)\in \mathcal{S}}l(y_t(\|\varphi_h-\varphi_i\|^2_\Mb-\|\varphi_h-\varphi_j\|_\Mb^2)) 
\end{aligned}
\end{equation}
where $n=3|\mathcal{S}|$ and $\varphi_i\in \mathbb{R}^{n}$ denotes the KPCA representation of feature $\phi_i\in\mathcal{H}$ for the random set $\phi(\xb_1), \phi(\xb_2), \ldots \phi(\xb_{n})$. We refer to the quantity $\widehat{\overline{R}}_\mathcal{S}(\Mb)$ as the (finite dimensional) empirical risk of $\Mb$. We can express $\widehat{L}_0$ using the solution of (finite dimensional) empirical risk minimization with corresponding constraints. In Section \ref{sec:KPCA} we use known results to explain how to perform KPCA, how to calculate distances with finite dimensional vectors in KPCA and how to relate norm constraints over $L$ with finite dimensional metric $\Mb$. Then, in Section \ref{sec: learning M}, we provide the finite dimensional optimization with all constraints that is equivalent to (\ref{opt-P3}) and express $\widehat{L}_0$ from its solution.  

\subsection{Kernelized Principle Component Analysis (KPCA)} \label{sec:KPCA}
In this part, we explain how to perform kernelized PCA in a reproducing kernel Hilbert space (RKHS). Consider the set of items $\bm{x}_1, \bm{x}_2, \bm{x}_3 \ldots \bm{x}_{n} \in \mathbb{R}^d$ and corresponding features $\phi_1, \phi_2, \ldots \phi_{n}$. We assume that $\phi_i$'s are linearly independent. Recall that $\mathcal{S}_\mathcal{X}\subset\mathcal{H}$ represents the subspace spanned by $\{\phi(\xb_1), \phi(\xb_2), \ldots \phi(\xb_{n})\}$. Let $\psi_1, \psi_2, \ldots \psi_n$ be the n principal component directions in this space. We show how to efficiently compute projections onto this subspace using the idea of Kernelized Principle Component Analysis (KPCA). This is important as the principal components live in the possibly infinite dimensional space $\mathcal{H}$ making traditional optimization either intractable or impossible. The following procedure, which we summarize for completeness from a previous study \cite{chatpatanasiri2010new} can be used to compute the projection of any point $\bm{x} \in \mathbb{R}^d$ onto the principal component directions in time that is polynomial in $n=3|\mathcal{S}|$:
\begin{enumerate}
    \item Form the Gram matrix: $\Kb \in \mathbb{R}^{n\times n}$ such that $\Kb_{i,j}=k(\bm{x}_i,\bm{x}_j)$.
    \item Center the Gram matrix: $\overline{\Kb}=\Kb-\frac{1}{n}\mathbf{1}_{n\times n}\Kb-\frac{1}{n}\Kb\mathbf{1}_{n\times n}+\frac{1}{n^2}\mathbf{1}_{n\times n}\Kb\mathbf{1}_{n\times n}$, where $\mathbf{1}_{n\times n}$ is the n by n matrix of all ones.
    \item Compute all n eigenvectors of $\overline{\Kb}, \alpha_1, \ldots, \alpha_{n}$ and form matrix $\Ab = [\alpha_1, \ldots, \alpha_{n}]$. 
    \item For any $\xb \in \mathbb{R}^d$ and any principal component $\psi_j$ with eigenvector $\alpha_j$, we have that $\langle \phi(\xb), \psi_j \rangle_\mathcal{H}=\sum_{i=1}^{n}\alpha_{i,j}k(\xb, \xb_i)$.
    \item Therefore, for any $\xb \in \mathbb{R}^d$ we may represent $\phi(\xb)$ in terms of its projection onto $\psi_1, \ldots, \psi_{n}$ as
    \begin{equation*}
        \varphi(\xb)=\Ab^T[k(\xb, \xb_1), \ldots, k(\xb, \xb_{n})]^T
    \end{equation*}
\end{enumerate}
For the remainder, we will let $\varphi_i\in \mathbb{R}^{n}$ denote the KPCA representation of random feature $\phi_i\in\mathcal{H}$ for the set $\phi(\xb_1), \phi(\xb_2), \ldots \phi(\xb_{n})$. The following representer theorem demonstrates that we may instead use finite dimensional vectors $\varphi_1, \ldots, \varphi_{n}$ for the optimization without loss in performance for a given set $\phi(\xb_1), \phi(\xb_2), \ldots \phi(\xb_{n})$.

\begin{prop}{(Theorem 1 in \cite{chatpatanasiri2010new})}\label{prop:representer theorem}
    Let $\{\overline{\psi}_i\}_{i=1}^n$ be any set of points in $\mathcal{H}$ such that Span$\left(\{\overline{\psi}_i\}_{i=1}^n\right)=\mathcal{S}_\mathcal{X}$ and let $\mathcal{H}'$ be a Hilbert space such that $\mathcal{H}$ and $\mathcal{H}'$ are separable. For any objective function $f$, the optimization
    \begin{equation*}
        \min_L f\left(\{\langle L\phi_i, L\phi_j\rangle_{\mathcal{H}'}\}_{i,j\in [n]}\right)
    \end{equation*}
    such that $L : \mathcal{H} \rightarrow \mathcal{H}'$ is a bounded linear map, has the same optimal value as
    \begin{equation*}
        \min_{{L}'\in\mathbb{R}^{n\times n}} f\left(\{\overline{\psi}(\xb_i)^T{{L}'}^T{L}'\overline{\psi}(\xb_j)\}_{i,j\in [n]}\right)
    \end{equation*}
    where $\overline{\psi}(\xb)=[\langle\phi(\xb), \overline{\psi}_1 \rangle, \ldots, \langle\phi(\xb), \overline{\psi}_n \rangle]^T\in \mathbb{R}^n$.
\end{prop}

\textbf{Calculating Kernelized Mahalanobis Distances using KPCA:}
Proposition \ref{prop:representer theorem} provides that one can learn $\widehat{L}_0$ using the KPCA representations of $\bm{x}_1, \bm{x}_2\ldots \bm{x}_n$. To be precise, given a linear map $L : \mathcal{H}\rightarrow \mathcal{H}$, we may expand the distance $\|L\phi_i-L\phi_j\|^2=\langle L\phi_i,L\phi_i\rangle-2\langle L\phi_i,L\phi_j\rangle+\langle L\phi_j,L\phi_j\rangle$. Let $\Ab$ be as defined in kernelized PCA and $\Phi:=[\phi_1, \phi_2, \ldots \phi_n ]$ denote the matrix whose columns are $\phi_i$’s. As the $\phi_i$’s are linearly independent, $\Phi$ is full rank\footnote{In the case where the $\phi_i$’s are not linearly independent and $\Phi$ is no longer full rank, KPCA can be modified by projecting onto the $k < n$ eigenvectors corresponding to the nonzero eigenvalues.}. For any $\phi_k$ within the set $\{\phi_1, \phi_2, \ldots \phi_n\}$, we have $L\phi_k=\Ub\Ab^T\Phi^T\phi_k$ for a linear map $\Ub$ from $\mathbb{R}^n$ to $\mathcal{H}$. Additionally, by definition of the kernel function $k(\cdot, \cdot)$, $\Phi^T\phi(\xb_k)=[k(\xb_k,\xb_1),\ldots,k(\xb_k,\xb_n)]^T$. Hence,
\begin{eqnarray*}
    \|L\phi_i-L\phi_j\|_\mathcal{H}^2 &=& \langle\Ub\varphi_i,\Ub\varphi_i\rangle-2\langle\Ub\varphi_i,\Ub\varphi_j\rangle+\langle\Ub\varphi_j,\Ub\varphi_j\rangle \nonumber
    \\ &=& \|\Ub\varphi_i-\Ub\varphi_j\|^2 \nonumber
    \\ &=&\|\varphi_i-\varphi_j\|_\Mb^2 \label{M for L}
\end{eqnarray*}
for $\varphi_i \in \mathbf{R}^n$ defined by kernelized PCA on $\phi_1, \phi_2, \ldots \phi_n$, and $\Mb = \Ub^T\Ub \in \mathbb{R}^{n\times n}$. Therefore, we may use kernelized PCA to efficiently compute distances in $\mathbb{R}^n$ as opposed to in $\mathcal{H}$ for a given set $\phi_1, \phi_2, \ldots \phi_n$.

\textbf{Relating norms in $\mathcal{H}$ and $\mathbb{R}^n$:}
Above lines demonstrate how, for a given $L$, we may find a specific $\Mb$ that defines a metric on $\mathbb{R}^n$ which computes distances between points in $\mathcal{S}_\mathcal{X}\subset\mathcal{H}$ equally to $L$ using the KPCA basis for $\mathcal{S}_\mathcal{X}$.

We consider Schatten norm constraints on $L$ to rigorously define model classes for $L$ in (\ref{opt-P1}), (\ref{opt-P2}) and (\ref{opt-P3}). Hence, it is necessary to relate the Schatten norms of $L$ to Schatten norms of $\Mb$ so that constraints placed on $L$ in (\ref{opt-P3}) are comparable to those placed on $\Mb$ in $\mathbb{R}^n$. Following Lemma relate these norms.
\begin{lemma}\label{lem:relating norms}
    Let $\phi_1, \ldots, \phi_n$ be a set of features corresponding to a random set of triplets and $\mathcal{S}_\mathcal{X}$ is the span of feature points. For any $\phi_\xb\in \mathcal{S}_\mathcal{X}$ and $L: \mathcal{H}\rightarrow \mathcal{H}$, there exists a semidefinite matrix $\Mb \in \mathbb{R}^{n\times n}$ such that $\|L\phi_\xb\|_\mathcal{H}=\|\varphi_\xb\|_\Mb$ and $\|\mathcal{P}^\dagger_{\mathcal{S}_\mathcal{X}}L^\dagger L\mathcal{P}_{\mathcal{S}_\mathcal{X}}\|_{S_p}=\|\Mb\|_p$, $\forall p$.
\end{lemma}
Now, we can set up a finite dimensional optimization problem to learn a finite dimensional metric $\Mb$ that will enable us to find $\widehat{L}_0$.

\subsection{Learning Kernelized Metrics in Practice}\label{sec: learning M}
We define following finite dimensional constrained convex program to learn a kernelized Mahalanobis metric from a random set of triplets $\mathcal{S}$:
\begin{equation}
\begin{aligned}
\min_{\Mb\succeq 0} \quad & \widehat{\overline{R}}_\mathcal{S}(\Mb)\\
\textrm{s.t.} \quad  
  &  \|\Mb\|_F \leq \lambda_F    \\
\end{aligned}\tag{P4}\label{opt-P4}
\end{equation}
where $\Mb \succeq 0$ denotes that $\Mb$ is positive semidefinite and the condition on the norm prevents overfitting as in (\ref{opt-P1}), (\ref{opt-P2}) and (\ref{opt-P3}). Let $\widehat{\Mb}$ denote an optimal solution to (\ref{opt-P4}) referred as the empirical risk minimizer. Likewise, if we instead consider $\|\mathcal{P}^\dagger_{\mathcal{S}_\mathcal{X}}L^\dagger L\mathcal{P}_{\mathcal{S}_\mathcal{X}}\|_{S_2}\leq\lambda_*$, this is corresponding to $\|\Mb\|_*\leq\lambda_*$ where $\|\cdot\|_*$ denotes the nuclear norm. In this setting, we may likewise solve for $\widehat{\Mb}$ satisfying this constraint instead. Below, Proposition \ref{prop:operationalizing} presents the relation between $(\ref{opt-P3})$ and $(\ref{opt-P4})$. Then, we show how to obtain $\widehat{L}_0$ from the finite dimensional solution. 
\begin{prop}\label{prop:operationalizing}
    Optimization problems (\ref{opt-P4}) and (\ref{opt-P3}) are equivalent. Solving $(\ref{opt-P4})$ is equal to learning $\widehat{L}_0$. Likewise, $\widehat{L}_0$ can be considered as the Hilbert space counterpart of finite dimensional space operator $\widehat{\Mb}$. Furthermore, let $\Psi_1,\ldots,\Psi_n \in \mathcal{H}$ be KPCA directions for the span $\mathcal{S}_{\mathcal{X}}$. We can write $\widehat{L}_0$ as
\begin{eqnarray}
\widehat{L}_0:\widehat{L}_0\phi_x =    \sum_{i=1}^n\sum_{j=1}^nw_{i,j}\Psi_i\otimes \Psi_j \mathcal{P}_{\mathcal{S}_{\mathcal{X}}}\phi_x \label{L0 def}
\end{eqnarray}
where $\Psi_i\otimes \Psi_j \phi_x = \langle \Psi_j, \phi_x\rangle_\mathcal{H}\Psi_i$ and $\Wb= \text{Chol}(\widehat{\Mb})$ such that $\Wb\Wb^T=\widehat{\Mb}$, i.e., $\Wb$ is from Cholesky decomposition of $\widehat{\Mb}$.
\end{prop}
Proposition \ref{prop:operationalizing} allows us to operationalize (\ref{opt-P3}) with a finite dimensional convex optimization problem and express $\widehat{L}_0$ from its solution.
\section{Experimental Results}\label{sec:experimental results}
In this part, we present simulations and experiments on real datasets to validate our theoretical results. Table \ref{tab:kernels} presents the kernel functions and the kernel parameters that we used in simulations and experiments. In all of our simulations and experiments, we use CVXPY \cite{diamond2016cvxpy, agrawal2018rewriting} and MOSEK \cite{mosek} to solve the convex program (\ref{opt-P4}). We use the nuclear norm constraint for $\Mb$ in (\ref{opt-P4}). 

\begin{table}[h]
\centering
\begin{tabular}{@{}lll@{}}
\toprule
Kernel     & Formula                                          & Parameter            \\ \midrule
Linear     & $k(x, y) = x^\top y$                             & N/A                  \\
Gaussian   & $k(x, y) = e^{\frac{-\|x - y\|^2_2}{2\sigma^2}}$ & $\sigma$             \\
Sigmoid    & $k(x, y) = \tanh{(c + \alpha x^\top y )}$        & $c, \alpha$          \\
Polynomial & $k(x, y) = (c + x^\top y)^p$     & $c, p$ \\
Laplacian  & $k(x, y) = e^{\alpha \|x - y\|_1}$               & $\alpha$             \\ \bottomrule
\end{tabular}
\caption{List of kernel functions and parameters used in our simulations and experiments.}
\label{tab:kernels}
\end{table}

To apply these kernel functions efficiently, especially on large datasets, we consider the computational complexity of the Kernelized Principal Component Analysis (KPCA) operation, which is ${O}(n^3)$, where $n$ is the number of items used in queries. To mitigate this cost, one can adapt low-rank approximations of the Gram matrix (Nyström method \cite{reinhardt2012analysis, williams1998prediction}) by randomly sampling $m \ll n$ items from $n$. The Nyström KPCA method \cite{williams2000using} has a complexity of ${O}(nm^2)$. Another approach, the randomly pivoted Cholesky algorithm \cite{chen2025randomly}, requires only ${O}(k^2n)$ kernel evaluations for a rank-$k$ approximation. In our work, we leverage the Nyström KPCA \cite{williams2000using} with $m=500$ to efficiently approximate the Gram matrix.
 
\textbf{Generating Noisy Labels for a Known Distance Function:} We assume an explicit link function $f(\cdot)$, where $f(\cdot)$ generates noisy labels for each triplet following that $y_t=-1$ with probability $p_t$ as a noisy indication of $\text{sign}(d_L^2 (\xb_h, \xb_i)-d_L^2 (\xb_i, \xb_j))$, where
\begin{eqnarray*}
    p_t=f\left(d_L^2 (\xb_h, \xb_i)-d_L^2 (\xb_h, \xb_j)\right).
\end{eqnarray*}
We use $f(x) = {1}/{(1 + e^{\rho x})}$ as the link function, where the parameter $\rho$ controls the noise level.

\textbf{Unseen Triplets: }To evaluate the performance of $\widehat{\Mb}$ for unseen triplets, we, first find $\varphi_{n+1}=\Ab^T[k(\xb_{n+1}, \xb_1), \ldots, k(\xb_{n+1}, \xb_{n})]^T$ for each new point $\bx_{n+1}$ seen in the test set using kernel function $k(x,y)$, where $\Ab$ is from KPCA procedure (see Section \ref{sec:KPCA}). This corresponds to finding the projections of new points to the span of $\phi_1\ldots \phi_n$. Then, we can estimate the label for an unseen triplet using new (finite) representations $\varphi_{n+1}$'s and $\widehat{\Mb}$.

\textbf{Computing Infrastructure: }Our code is designed to run on a personal laptop. Experiments and simulations reported in this paper were conducted on a MacBook Pro with M3 Max CPU with 48GB of RAM.

\begin{figure}[h]
    \centering
    \includegraphics[width=0.7\linewidth]{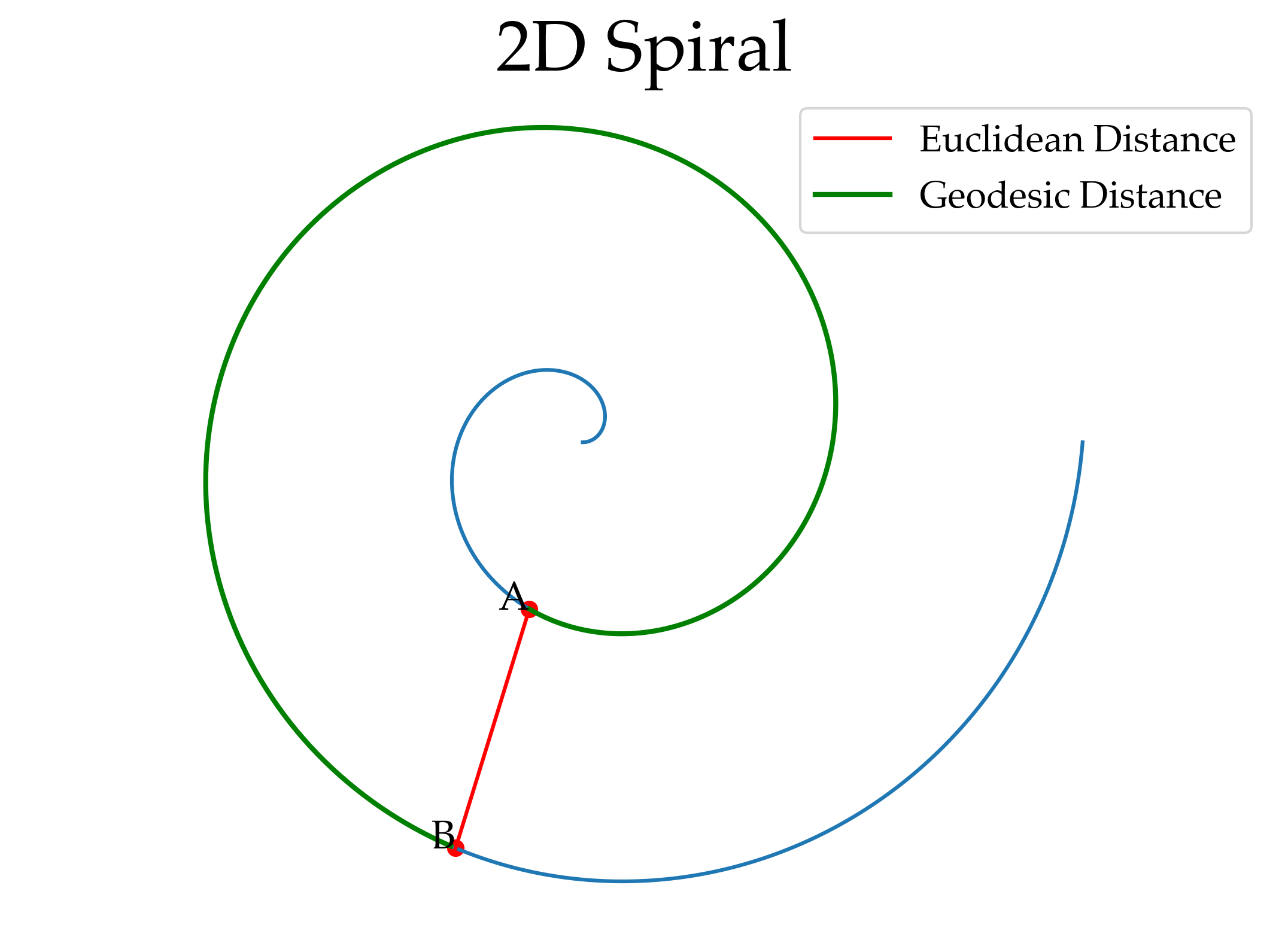}
    \caption{A 2D spiral. We sample triplets uniformly along this curve. The geodesic distance between point $A$ and $B$ is the length of the green curve, whereas the Euclidean distance between the two points is the length of the red line.}
    \label{fig:2dspiral}
\end{figure}

\subsection{Simulations: Spiral with Geodesic Distance}\label{sec:simulations}
\begin{figure}[h]
    \centering
    \includegraphics[width=\linewidth]{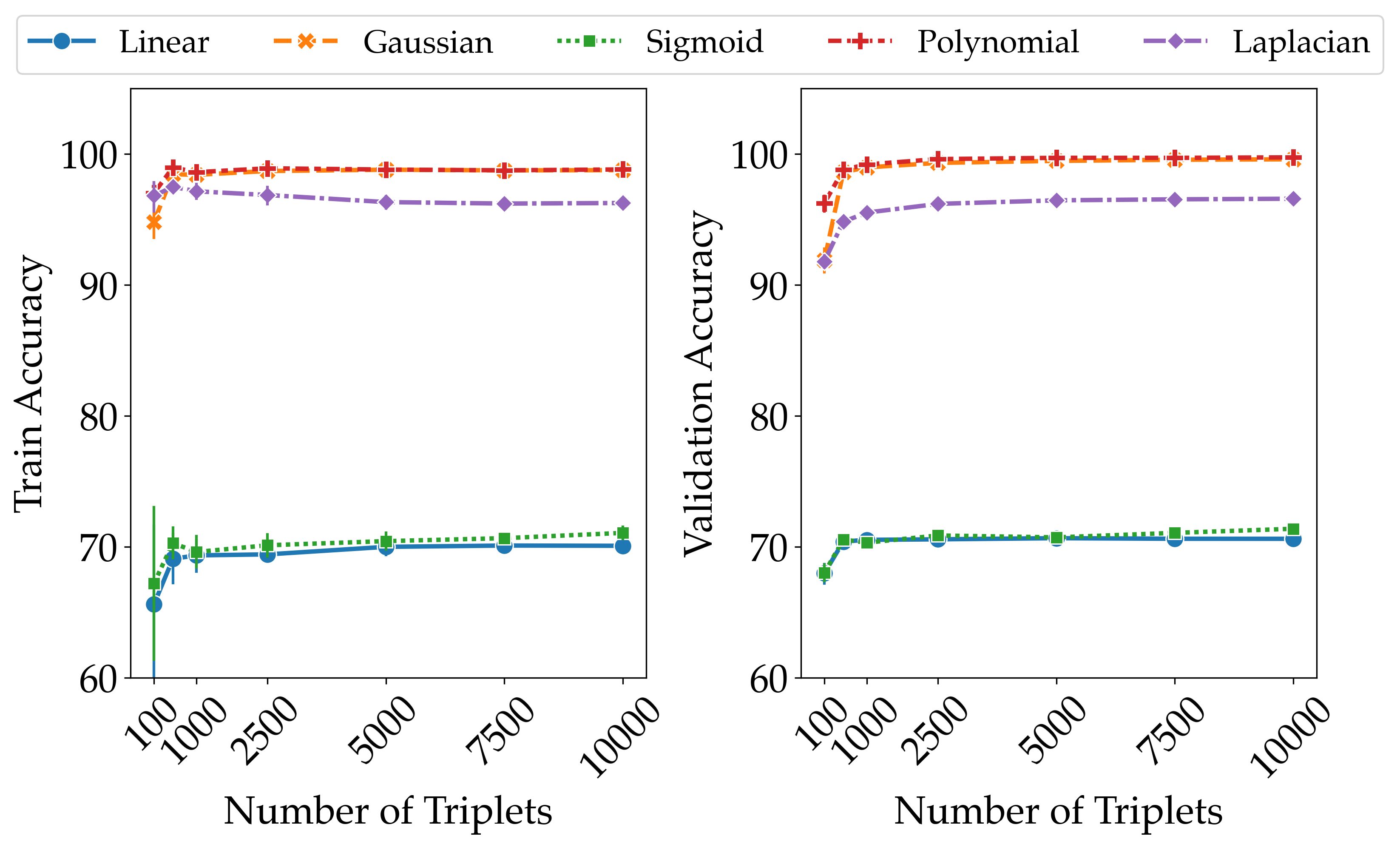}
    \caption{Performance of various kernels in the 2D spiral setting. For the Gaussian kernel, we use $\sigma=2$; sigmoid kernel, $c=1, \alpha=1$; polynomial kernel, $c=1, p=2$, Laplacian kernel, $\alpha=1$. For the link function $f$, we use $\rho=30$ to set the noise level around 0.01. We repeat each run 50 times.}
    \label{fig:spiral}
\end{figure}

We first consider a spiral shape in 2D. We generate triplets uniformly along the spiral. We assume the true distance function is the geodesic distance (see Figure \ref{fig:2dspiral}) along the 2D curve. We provide train and test accuracy for different kernel functions with varying number of triplets. Figure \ref{fig:spiral} illustrates the performance of various kernels. We observe that polynomial, Gaussian, and Laplacian kernels outperform linear and sigmoid kernels.

The test accuracy increases as we have more triplets for training in Figure \ref{fig:simulation}. We also observe that, as the number of triplets increases, the train and test accuracy gets close, consistent with our analysis in Theorems \ref{thm:generalization_error_withbounded_Fro_norm} and \ref{thm:generalization_error_withbounded_Nuclear_norm}. Recall that excess risk decreases with more triplets according to Theorems \ref{thm:generalization_error_withbounded_Fro_norm} and \ref{thm:generalization_error_withbounded_Nuclear_norm}.

\subsection{Simulations: Gaussian Kernel Map}
We assume that we have access to a feature map $\phi$ such that $\langle \phi(\bx_i), \phi(\bx_j)\rangle=k(\bx_i, \bx_j)$ with a Gaussian kernel function $k: \mathbb{R}^d\times \mathbb{R}^d \rightarrow \mathbb{R}^1$, where $\sigma=1$. We first want to generate a ground truth linear functional $L^*:\mathcal{H} \rightarrow \mathcal{H}$ that lies on an $r$-dimensional manifold. 

\textbf{Preliminary: }Note that Riesz's Representation Theorem allows us to represent the linear functional $L^*$ as follows:
\begin{eqnarray*}   L^*\phi=\sum_{k=1}^\infty\langle \phi, \tau_k\rangle_\mathcal{H} \mathbf{e}_k.
\end{eqnarray*}
Given that $L^*$ lies on an $r$-dimensional manifold, each $\tau_k$ can be written as $\sum_{j=1}^rv_{k,j}\psi_j$, where $\{\psi_1, \ldots, \psi_r\}$ is a set of features that span an $r$-dimensional manifold. Therefore, for any $\phi_i, \phi_j$,
\begin{eqnarray}
    \langle L\phi_i, L\phi_j\rangle_\mathcal{H} &=& \sum_{k=1}^\infty\langle \phi_i, \tau_k \rangle_\mathcal{H} \langle \phi_j, \tau_k \rangle_\mathcal{H} \nonumber
    \\ &=& \sum_{a=1}^r\sum_{b=1}^r \left(\sum_{k=1}^\infty v_{k,a}v_{k,b}\right)\langle \phi_i, \psi_a \rangle_\mathcal{H} \langle \phi_j, \psi_b \rangle_\mathcal{H}, 
    \label{Riesz_representationapp}
\end{eqnarray}
where $\Gb_{a,b}=\left(\sum_{k=1}^\infty v_{k,a}v_{k,b}\right)$. Each entry of $\Gb$ is an inner product in $\ell_2$, so $\Gb$ is a positive semidefinite matrix. Our target is to sample a set of features in $\mathcal{H}$ that spans an $r_0$-dimensional manifold, where $r_0=\max{(r)}$ and generate a random psd matrix $\Gb$ to define $L^*$. Inspired by the simulation setup of a previous work \cite{mason2017learning} on the linear metric learning problem, we define $\Gb$ as $\Gb=\frac{r_0}{\sqrt{r}}\Ub\Ub^T$ to ensure that the average magnitude of entries remains constant regardless of $r$ and $r_0$, where $\Ub\in \mathbb{R}^{r_0\times r}$ is a random orthogonal matrix. This procedure results in a linear functional $L^*$ lying on an $r$-dimensional manifold.  

\textbf{Linear Functional $L^*$: }We sample a set $\{z_1 \ldots z_r\}$, where each $z_i \sim \mathcal{N}(\textbf{0}_d, \frac{1}{d} I_d)$. Then, consider a kernel map $\phi(\cdot)$ such that $\langle \phi(z_i), \phi(z_j) \rangle=k(z_i, z_j)$. We generate corresponding features using this kernel map, where the set of features $\{\phi(z_1)\ldots \phi(z_r)\}$ span an $r$-dimensional manifold in $\mathcal{H}$ and define $\psi_i=\phi(z_i)$. We also generate a random psd matrix $\Gb_{r\times r}$. Finally, we have an explicit formula for $L^*$ based on (\ref{Riesz_representationapp}). Now, we can express inner product $\langle L\phi_i, L\phi_j\rangle_\mathcal{H}$ in terms of known parameters: 
\begin{eqnarray}
     \langle L\phi_i, L\phi_j\rangle_\mathcal{H} = [k(\bx_i, z_1), \ldots, k(\bx_i, z_r)]\Gb [k(\bx_j, z_1), \ldots, k(\bx_j, z_r)]^T, \label{inner product_app}
\end{eqnarray}
where $\langle \phi_i, \psi_a\rangle_\mathcal{H}=k(\bx_i, z_a)$  and $\phi_i=\phi(\bx_i)$. We can easily find the difference of distances for triplet comparisons using (\ref{inner product_app}), since we have 
$$\|L\phi(\xb_h)-L\phi(\xb_i)\|_\mathcal{H}^2=\langle L\phi_h,L\phi_h\rangle_\mathcal{H}-2\langle L\phi_h,L\phi_i\rangle_\mathcal{H}+\langle L\phi_i,L\phi_i\rangle_\mathcal{H}.$$

\textbf{Triplet Generation: } We randomly sample triples $\{\bx_h,\bx_i,\bx_j\}$ where $\bx_i\sim \mathcal{N}(\textbf{0}_d, \frac{1}{d}I_d)$. Then, we can numerically calculate the difference of distances using (\ref{inner product_app}) and generate noisy responses for triplets with a link function as mentioned in Section \ref{sec:experimental results}.   

\textbf{Accuracy: }We generate another set of random triplets. We can numerically find the true label corresponding to each triplet using $L^*$. Finally, we compare the true labels with the estimated labels to find accuracy. 

\begin{figure}[h]
    \centering
    \includegraphics[width=0.5\linewidth]{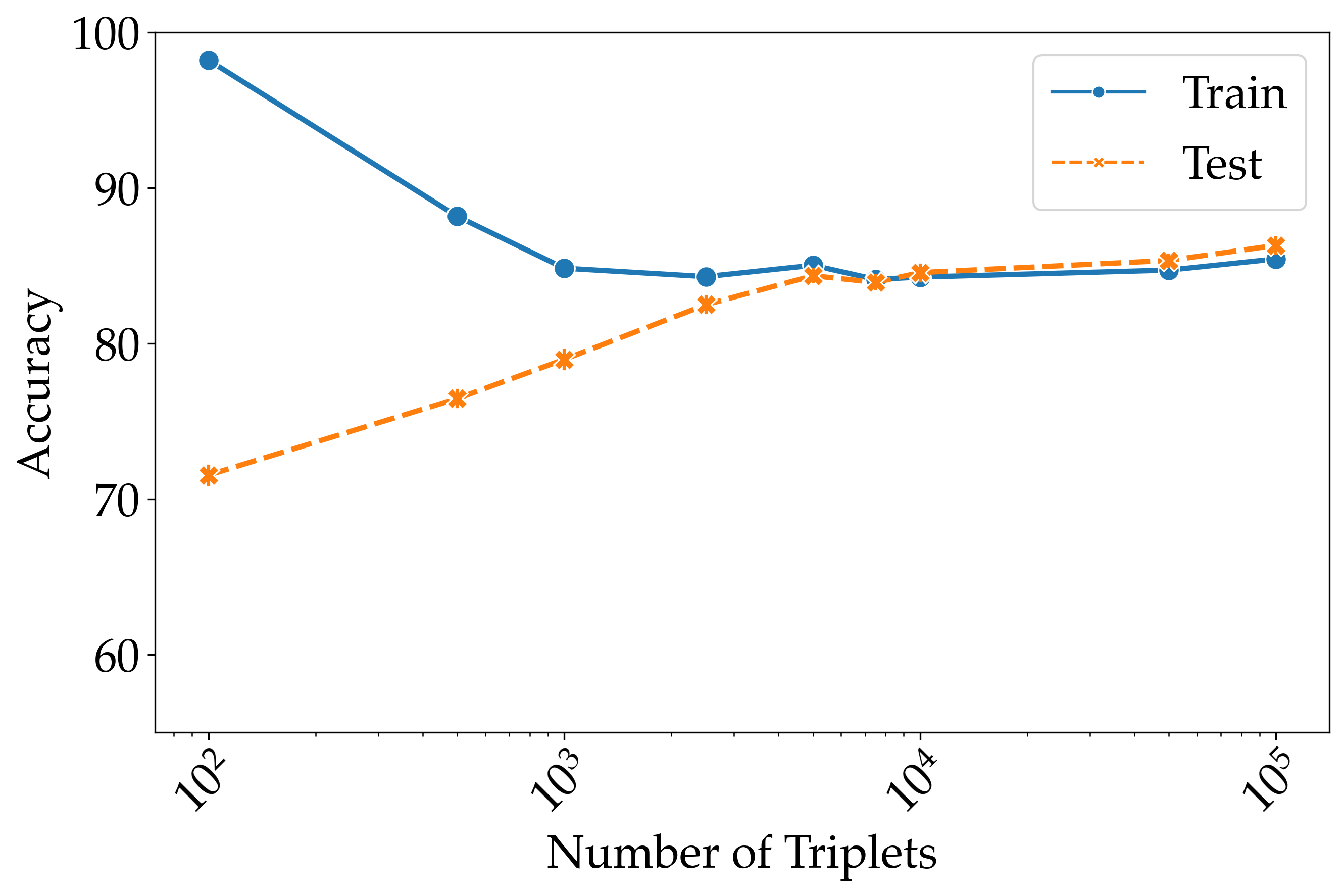}
    \caption{Train and test accuracy of Gaussian kernel. Here, we use $\sigma=1$. For the link function $f$, we use $\rho=1000$ to set the noise level around 0.05. We repeat each run 50 times.}
    \label{fig:simulation}
\end{figure}

We provide simulations with a Gaussian kernel with $\sigma=1$, where $k(x, y) = e^{\frac{-\|x - y\|^2_2}{2\sigma^2}}$. In Figure \ref{fig:simulation}, we provide a result with a Gaussian kernel for $r=2$. The test accuracy increases as we have more triplets for training for the task of learning a metric that lies on a $2$-dimensional manifold. The test accuracy increases as we have more triplets for training in Figure \ref{fig:simulation}. We also observe that, as the number of triplets increases, the train and test accuracy gets close, consistent with our analysis in Theorems \ref{thm:generalization_error_withbounded_Fro_norm} and \ref{thm:generalization_error_withbounded_Nuclear_norm}. Recall that excess risk decreases with more triplets according to Theorems \ref{thm:generalization_error_withbounded_Fro_norm} and \ref{thm:generalization_error_withbounded_Nuclear_norm}.

\begin{figure}[H]
\centering
\begin{minipage}{.5\textwidth}
  \centering
  \includegraphics[width=1\linewidth]{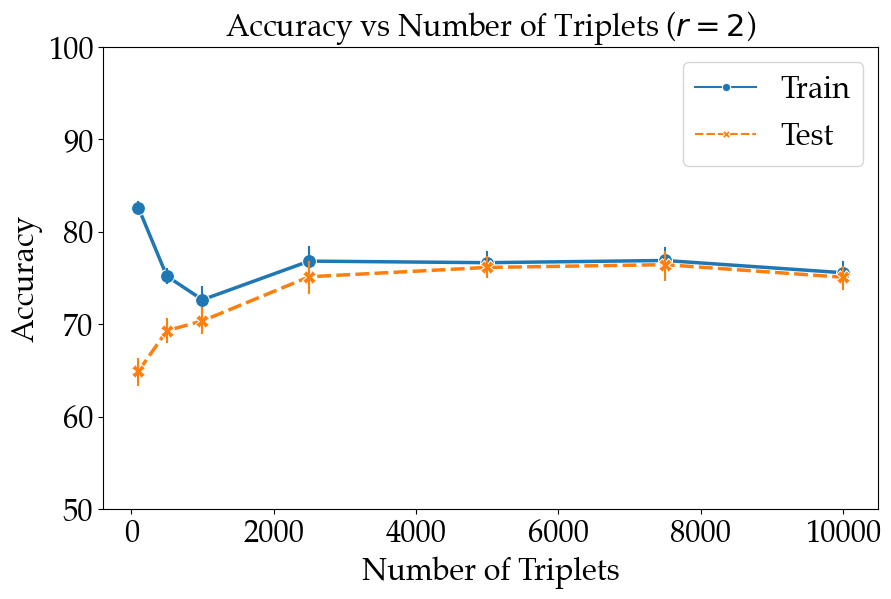}
\end{minipage}%
\begin{minipage}{.5\textwidth}
  \centering
  \includegraphics[width=1\linewidth]{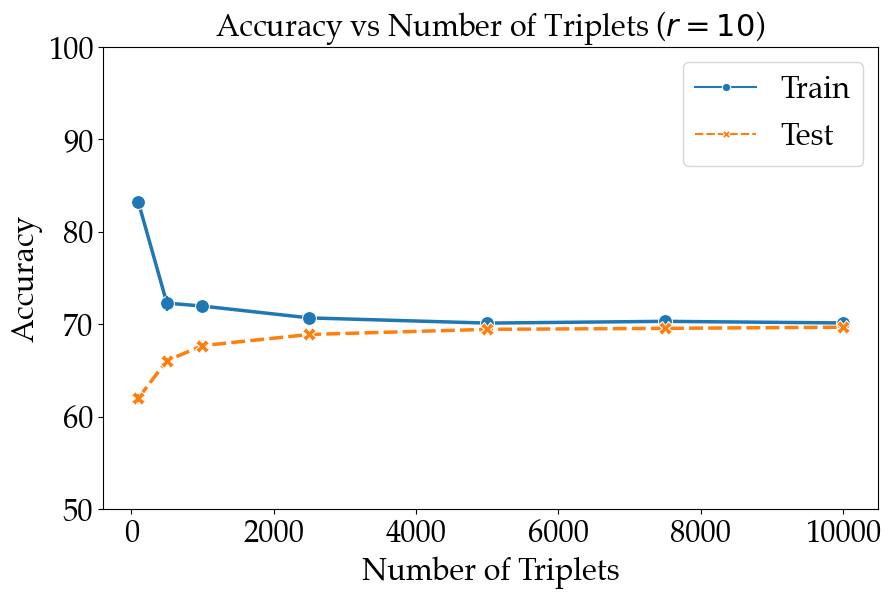}
\end{minipage}
\caption{Train and test accuracy for noiseless setting with 50 repetitions varying number of triplets (100, 500, 1000, 2500, 5000, 10000), where $r=2$ (left) and $r=10$ (right).}
\label{fig:r10noiseless}
\end{figure}
Figure \ref{fig:r10noiseless} shows that test accuracy increases when the triplet set gets larger. As a result, the learned metric generalizes better. For example, we observe that, to obtain the same accuracy of $70\%$, $\sim 1000$ triplets are sufficient when rank is 2, whereas the triplets needed when rank is 10 is $\sim 5000$.

\begin{figure}[H]
\centering
\begin{minipage}{.5\textwidth}
    \centering
    \includegraphics[width=\linewidth]{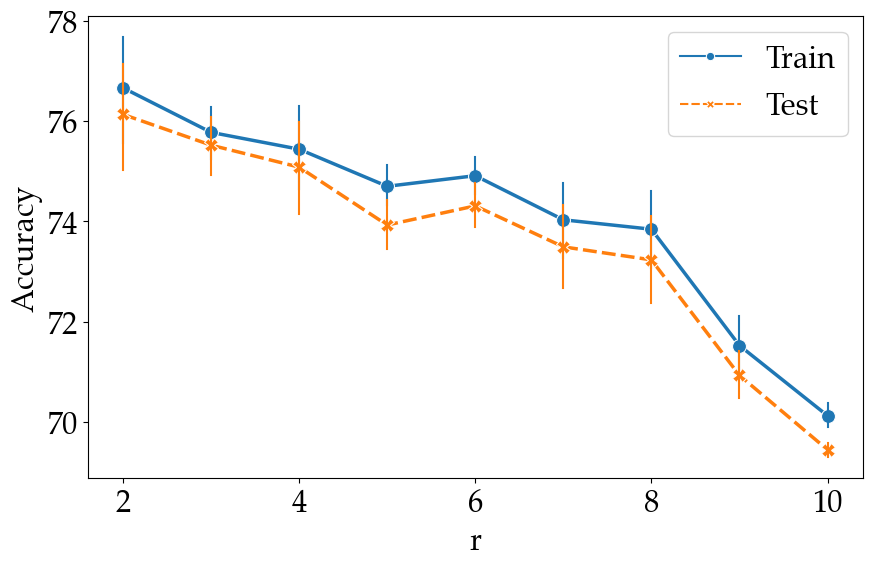}
\end{minipage}%
\begin{minipage}{.5\textwidth}
    \centering
    \includegraphics[width=\linewidth]{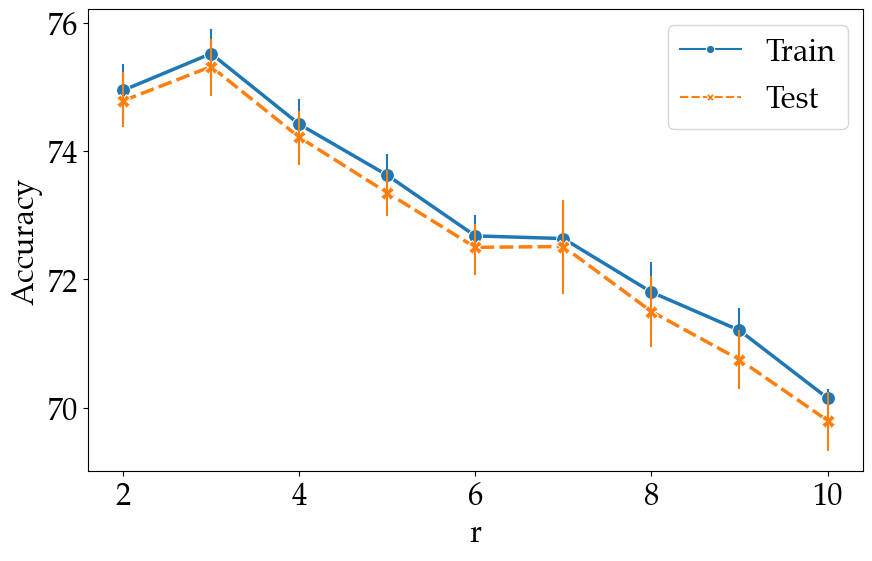}
\end{minipage}
\caption{Train and test accuracy using a fixed number of triplets and varying $r$ from 2 to 10, with 50 repetitions. Noiseless setting on the left with 5000 triplets and the ratio of noisy responses is approximately $5\%$ on the right with 10000 triplets.}
\label{fig:noiseless_r2to10}
\end{figure}

\begin{figure}[h]
\centering
\begin{minipage}{.5\textwidth}
  \centering
  \includegraphics[width=1\linewidth]{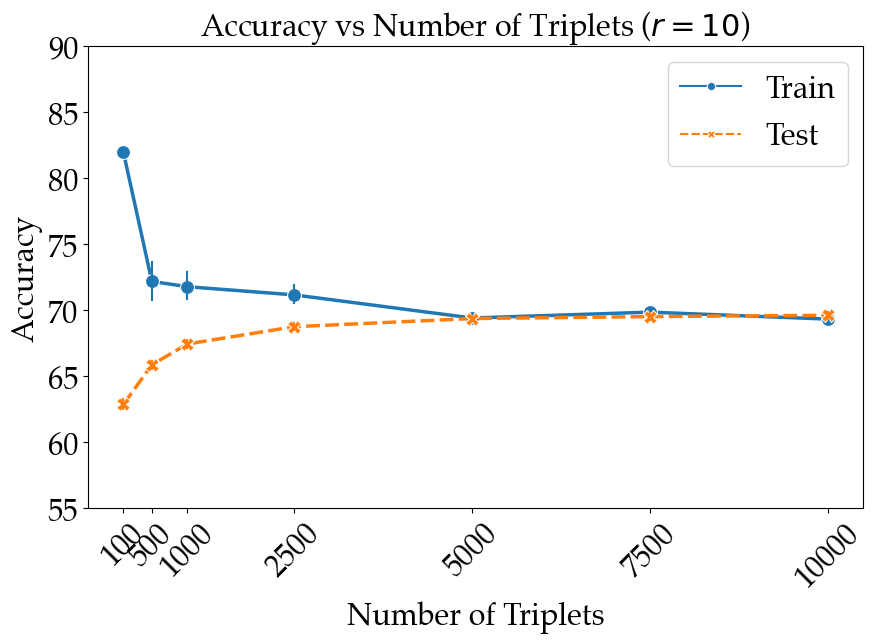}
\end{minipage}%
\begin{minipage}{.5\textwidth}
  \centering
  \includegraphics[width=1\linewidth]{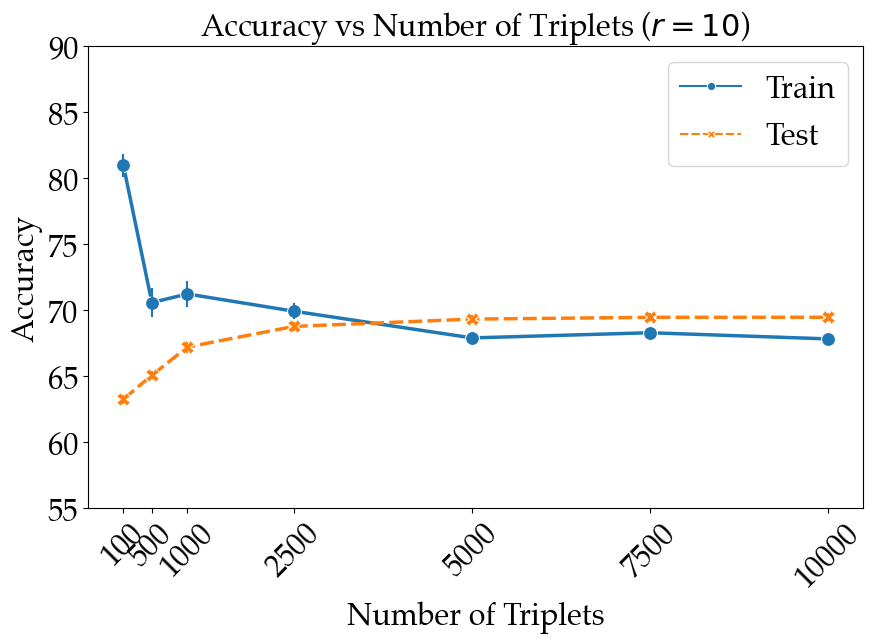}
\end{minipage}
\caption{Train and test accuracy for noisy setting and $r=10$ with 20 repetitions varying number of triplets, where the ratio of noisy responses is approximately $5\%$ (left) and $10\%$ (right).}
  \label{fig:r10noise10}
\end{figure}

In Figure \ref{fig:noiseless_r2to10}, we observe that for a fixed number of triplets, the accuracy that can be obtained decreases as the rank $r$ increases, as captured by our analysis, where $L^*$ lies on an $r$-dimensional manifold. The task of learning a kernelized metric becomes more complex as $r$ increases. We provide simulation results with both noiseless and noisy responses. Finally, Figure \ref{fig:r10noise10} shows the accuracy for varying numbers of triplets at noise levels of $5\%$ and $10\%$.

\subsection{Empirical Evaluation: Food-100 Dataset}

\begin{figure}[h]
    \centering
    \includegraphics[width=\linewidth]{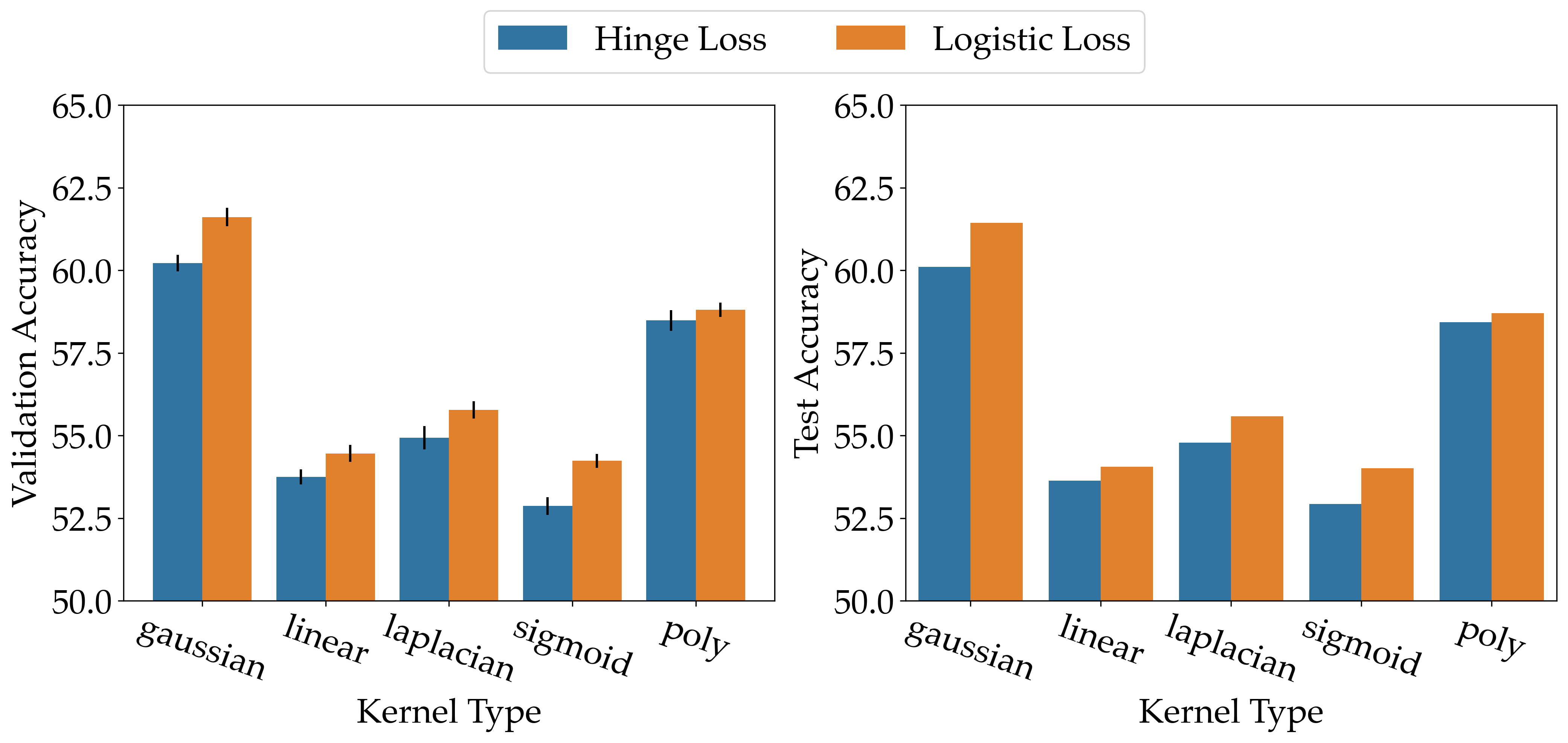}
    \caption{Performance of various kernels under the Food-100 dataset \cite{wilber2014cost}. For the Gaussian kernel, we use $\sigma=2$; sigmoid kernel, $c=1, \alpha=0.01$; polynomial kernel, $c=1, p=2$, Laplacian kernel, $\alpha=1$. We repeat the validation 20 times.}
    \label{fig:experiment}
\end{figure}

The Food-100 dataset \cite{wilber2014cost} consists of 100 food items and approximately 190,000 triplets based on human responses (See Figure \ref{fig:triplets} for example images from the dataset). We divide this dataset by items to ensure that the model does not encounter some items in the test and validation sets during the training phase. We obtain embeddings for each item in Food-100 dataset using the embeddings from the antepenultimate layer of AlexNet \cite{krizhevsky2012imagenet}, pretrained on ImageNet \cite{deng2009imagenet}. We, then, project them to a 2D space using PaCMAP \cite{JMLR:v22:20-1061}. Figure \ref{fig:experiment} shows the performance of different kernels, among which the Gaussian kernel performs the best.

Theorems \ref{thm:generalization_error_withbounded_Fro_norm} and \ref{thm:generalization_error_withbounded_Nuclear_norm} provide bounds for excess risk. Therefore, our analysis allows us to bound the difference between the true risk and the empirical risk for any kernel choice. Experiments with different kernels demonstrate that train and test accuracies are close, indicating that the empirical risk approximates the true risk well. Choice of kernel has an effect on the true risk and therefore affects the risk achievable by the learned metric. This is reflected in the difference in test accuracies across different kernels. Since there is no way of knowing what the true risk is, cross-validation is an appropriate method for selecting the optimal kernel for the dataset at hand.

\textbf{Splitting triplets in the dataset: }The Food-100 dataset consists of carefully selected 100 food items, where each image has only one food. Answers to   190,376 triplets are collected from Amazon Mechanical Turk workers. Let $\mathcal{T}$ be the set of all triplets. For each iteration, we randomly select 20 items and call them $\mathcal{X}_\text{unseen}$. Then, we define a triplet set $\mathcal{T}_\text{unseen}$ from $\mathcal{X}_\text{unseen}$ as follows:
\begin{equation*}
    \mathcal{T}_\text{unseen} := \{ \{x_h, x_i, x_j\} : x_h \in \mathcal{X}_\text{unseen} \text{ or } x_i \in \mathcal{X}_\text{unseen} \text{ or } x_j \in \mathcal{X}_\text{unseen}\}.
\end{equation*}
Next, we uniformly sample triplets for the training set $\mathcal{T}_\text{train}$ from the set $\mathcal{T}\setminus \mathcal{T}_\text{unseen}$ to guarantee that there exist unseen items in $\mathcal{T}_\text{train}$. Finally, we uniformly sample triplets for the test set $\mathcal{T}_\text{test}$ from the set of all triplets $\mathcal{T}$. We apply the same splitting strategy on the $\mathcal{T}_\text{train}$ set to further split it to different training and validation part 20 times. We report the mean and standard deviation of the validation accuracies on these 20 validation parts.

\textbf{Choice of Parameters for Kernel Function: }
We conducted a parameter search on the validation set in the following range:
\begin{itemize}
    \item $\sigma: 0.01, 0.1, 1, 10$
    \item $\alpha: 0.01, 0.1, 1$
    \item $p: 2, 5, 7, 10$
\end{itemize}
 Our results show the best test accuracy values based on this search. 

\section{Conclusions and Future Work} \label{sec:conclusion}
When undertaking the task of developing a theoretical understanding of triplet based nonlinear metric learning methods, the first natural setting to consider is kernelized metric learning. To the best of our knowledge, there are no generalization results that analyze sample complexity for kernelized metric learning via triplet comparisons in the literature prior to our work. The theoretical foundations for metric learning via triplet queries are currently limited to linear settings, e.g., \cite{mason2017learning}, which provide generalization results for the linear setting when the set of items being queried is fixed and the number of items $n\gg d$ (See Appendix \ref{appendix:discussion} for further explanation). Therefore, our work fills an important gap in the literature. 

We provide a theoretical analysis for the kernelized metric learning problem. We provide novel generalization and sample complexity bounds. Developing an understanding of other nonlinear metric learning approaches, especially neural networks based approaches would be interesting for future research directions. That said, kernelized approaches are preferred in areas where interpretability and explainability are crucial, especially when they also perform nearly as well as other methods \cite{radhakrishnan2023transfer}. Therefore, understanding kernelized settings is also of value beyond theoretical pursuit towards understanding a broader set of nonlinear approaches. 
\section{Acknowledgements}
This work was partially supported by NSF grants NCS-FO 2219903 and NSF CAREER Award CCF 2238876.

\bibliographystyle{alpha}
\bibliography{main.bib}

\newpage
\appendix

\section{Proofs}\label{sec:Proofs}
\subsection{Proof of Lemma \ref{lem:norm inequality}}
We first note that orthogonal projections are bounded linear operators. Therefore, $\mathcal{P}_{\mathcal{S}_\mathcal{X}}$'s are bounded and linear. One can easily show that compositions of bounded linear operators are also bounded and linear. Therefore, $L\mathcal{P}_{\mathcal{S}_\mathcal{X}}$ is bounded and linear for any orthogonal projection $\mathcal{P}_{\mathcal{S}_\mathcal{X}}$. Then, for $p\geq 1$, we have
\begin{eqnarray*}
    \|\mathcal{P}^\dagger_{\mathcal{S}_\mathcal{X}}L^\dagger L\mathcal{P}_{\mathcal{S}_\mathcal{X}}\|_{S_p} &\overset{a}{\leq}& \|\mathcal{P}_{\mathcal{S}_\mathcal{X}}^\dagger\|_{S_\infty} \|L^\dagger L\mathcal{P}_{\mathcal{S}_\mathcal{X}}\|_{S_p} 
    \\ &\overset{b}{\leq}& \|\mathcal{P}_{\mathcal{S}_\mathcal{X}}^\dagger\|_{S_\infty} \|L^\dagger L\|_{S_p} \|\mathcal{P}_{\mathcal{S}_\mathcal{X}}\|_{S_\infty}
    \\ &=&  \|L^\dagger L\|_{S_p}, 
\end{eqnarray*}
where $(a)$ and $(b)$ follow from Hölder's inequality. Note that $\|\mathcal{P}_{\mathcal{S}_\mathcal{X}}\|_{S_\infty}$ is the standard operator norm on $\mathcal{H}$, i.e., $\|\mathcal{P}_{\mathcal{S}_\mathcal{X}}\|_{S_\infty}= \max_{\|x\|\leq 1}\|\mathcal{P}_{\mathcal{S}_\mathcal{X}} x\|$. Since $\mathcal{P}_{\mathcal{S}_\mathcal{X}}$ is an orthogonal projection, we have $\|\mathcal{P}_{\mathcal{S}_\mathcal{X}}\|_{S_\infty}=\|\mathcal{P}_{\mathcal{S}_\mathcal{X}}^\dagger\|_{S_\infty}=1$.

\subsection{Proof of Proposition \ref{prop:opt2 and opt3}}
We rewrite $\widehat{L}$ and $\widehat{L}_0$ together with (\ref{opt-P2}) and (\ref{opt-P3}) below for the ease of readability.
\begin{equation}
\begin{aligned}
\widehat{L}:=\arg \min_{L} \quad  \widehat{R}_\mathcal{S}(L) \quad
\textrm{s.t.} \quad  \|L^\dagger L\|_{S_2}\leq \lambda_F    \\
\end{aligned} \tag{P2}
\end{equation}

\begin{equation}
\begin{aligned}
\widehat{L}_0:=\arg\min_{L} \quad & \widehat{R}_\mathcal{S}(L) \quad
\textrm{s.t.} \quad  \|\mathcal{P}_{\mathcal{S}_\mathcal{X}}^\dagger L^\dagger L\mathcal{P}_{\mathcal{S}_\mathcal{X}}\|_{S_2}\leq \lambda_F    \\
\end{aligned} \tag{P3}
\end{equation}
We have $\|\widehat{L}^\dagger\widehat{L}\|_{S_2}\leq \lambda_F$ by definition. From Lemma \ref{lem:norm inequality}, we also have 
\begin{eqnarray*}
   \|\mathcal{P}^\dagger_{\mathcal{S}_\mathcal{X}}\widehat{L}^\dagger \widehat{L}\mathcal{P}_{\mathcal{S}_\mathcal{X}}\|_{S_2} \leq  \|\widehat{L}^\dagger\widehat{L}\|_{S_2}.
\end{eqnarray*}
Thus, it holds that $\|\mathcal{P}^\dagger_{\mathcal{S}_\mathcal{X}}\widehat{L}^\dagger \widehat{L}\mathcal{P}_{\mathcal{S}_\mathcal{X}}\|_{S_2}\leq \lambda_F$.
Therefore, $\widehat{L}$ belongs to the feasible set of optimization problem (\ref{opt-P3}) and we can conclude that $\widehat{{R}}_\mathcal{S}(\widehat{L})$ is at least as small as $\widehat{{R}}_\mathcal{S}(\widehat{L}_0)$, i.e., 
\begin{eqnarray}
    \widehat{{R}}_\mathcal{S}(\widehat{L}_0) \leq \widehat{{R}}_\mathcal{S}(\widehat{L}). \label{emp of L0 smaller than L}
\end{eqnarray}
For the reverse inequality, note that $\widehat{L}_0 = \widehat{L}_0\mathcal{P}_{\mathcal{S}_\mathcal{X}}$. Therefore, $\|\mathcal{P}^\dagger_{\mathcal{S}_\mathcal{X}}\widehat{L}_0^\dagger \widehat{L}_0\mathcal{P}_{\mathcal{S}_\mathcal{X}}\|_{S_2}=\|\widehat{L}_0^\dagger \widehat{L}_0\|_{S_2}$ and $\widehat{L}_0$ belongs to the feasible set of (\ref{opt-P2}). Hence, 
\begin{eqnarray}
   \widehat{{R}}_\mathcal{S}(\widehat{L})\leq \widehat{{R}}_\mathcal{S}(\widehat{L}_0). \label{emp of L smaller than L0}
\end{eqnarray}
Based on (\ref{emp of L0 smaller than L}) and (\ref{emp of L smaller than L0}), we find that
\begin{eqnarray}
   \widehat{{R}}_\mathcal{S}(\widehat{L})= \widehat{{R}}_\mathcal{S}(\widehat{L}_0).
\end{eqnarray}
Furthermore, note that any ${L}$ within the feasible set of (\ref{opt-P2}) also belongs to the feasible set of (\ref{opt-P3}). Provided that ${L}\mathcal{P}_{\mathcal{S}_\mathcal{X}}=\widehat{L}_0$, we also conclude that $L$ is an optimal solution for (\ref{opt-P2}), since $\widehat{{R}}_\mathcal{S}({L}\mathcal{P}_{\mathcal{S}_\mathcal{X}})=\widehat{{R}}_\mathcal{S}(\widehat{L}_0)=\widehat{{R}}_\mathcal{S}(\widehat{L})$. 

\subsection{Proof of Theorem \ref{thm:generalization_error_withbounded_Fro_norm}}
From Proposition \ref{prop:opt2 and opt3}, we have 
\begin{eqnarray}
    \widehat{{R}}_\mathcal{S}(\widehat{L}_0) = \widehat{{R}}_\mathcal{S}(\widehat{L}). \label{eq:empirical risk of L0 and L}
\end{eqnarray}
Then, using standard Rademacher complexity bounding techniques, we can write following
\begin{eqnarray}
R(\widehat{L}_{0})-R(L^*)&=& R(\widehat{L}_{0})-\widehat{R}_\mathcal{S}(\widehat{L}_{0})+\widehat{R}_\mathcal{S}(\widehat{L}_{0})-\widehat{R}_\mathcal{S}(L^*)+\widehat{R}_\mathcal{S}(L^*)-R(L^*) \nonumber
    \\
    &\overset{a}{=}& R(\widehat{L}_{0})-\widehat{R}_\mathcal{S}(\widehat{L}_{0})+\widehat{R}_\mathcal{S}(\widehat{L})-\widehat{R}_\mathcal{S}(L^*)+\widehat{R}_\mathcal{S}(L^*)-R(L^*) \nonumber 
    \\ &\leq&
    2 \sup_{L}|\widehat{R}_\mathcal{S}({L})-R(L)| \nonumber
    \\ 
    &\leq& 2\mathbb{E}_{\mathcal{S}\sim D}[\sup_{L}|\widehat{R}_\mathcal{S}({L})-R(L)|] + \beta\sqrt{\frac{2\ln{2/\delta}}{|\mathcal{S}|}}\label{Rademacher last}
\end{eqnarray}
where $\beta:= \sup_{} |\ell (\|L\phi_h-L\phi_i\|_\mathcal{H}^2-\|L\phi_h-L\phi_j\|_\mathcal{H}^2)-\ell (\|L\phi_h'-L\phi_i'\|_\mathcal{H}^2-\|L\phi_h'-L\phi_j'\|_\mathcal{H}^2)|$ and $(a)$ is from (\ref{eq:empirical risk of L0 and L}). Note that $\beta\leq 12\alpha\lambda_F B^2$, since the difference of triplets is bounded by $6\lambda_F B^2$ (see Lemma \ref{lem:difference of distances}) and the loss is $\alpha-$Lipschitz. 

Now, using standard symmetrization and contraction lemmas, we may introduce $\epsilon_t \in \{-1, 1\}$'s, that are Rademacher random variables corresponding to each triplet $t$. Then, we have
\begin{eqnarray*}
    \mathbb{E}_{\mathcal{S}\sim \mathcal{D}}[\sup_{L}|\widehat{R}_\mathcal{S}({L})-R(L)|]
    \leq 
    \mathbb{E}_{\mathcal{S}\sim \mathcal{D}, \epsilon \sim \{\pm 1\} ^{|\mathcal{S}|}} \frac{2\alpha}{|\mathcal{S}|} \left[\sup_L \sum_{t\in \mathcal{S}}\epsilon_t(\|L\phi_h-L\phi_i\|_\mathcal{H}^2-\|L\phi_h-L\phi_j\|_\mathcal{H}^2))\right]
\end{eqnarray*}
The expression inside the expectation on the right hand side can be considered as a function of random triplets in $\mathcal{S}$. We focus on the expectation on the right hand side:
\begin{eqnarray}
    \mathbb{E}_{\mathcal{S}\sim \mathcal{D}, \epsilon \sim \{\pm 1\} ^{|\mathcal{S}|}} \left[\sup_L \sum_{t\in \mathcal{S}}\epsilon_t(\|L\phi_h-L\phi_i\|_\mathcal{H}^2-\|L\phi_h-L\phi_j\|_\mathcal{H}^2))\right]. \label{expectation}
\end{eqnarray}
Note that (\ref{expectation}) is finite, since the difference of triplets is bounded. Therefore, we can apply Fubini's Theorem, and write it as
\begin{eqnarray}
    \mathbb{E}_{\mathcal{S}}\left[\mathbb{E}_{\epsilon |\mathcal{S}} \left[\sup_L \sum_{t\in \mathcal{S}}\epsilon_t(\|L\phi_h-L\phi_i\|_\mathcal{H}^2-\|L\phi_h-L\phi_j\|_\mathcal{H}^2))\right] \right] \label{cond_exp}
\end{eqnarray}
where $\mathbb{E}_{\epsilon |\mathcal{S}}$ is the conditional expectation given $\mathcal{S}$. In (\ref{expectation}), we have a set of random triplets with corresponding random features $\phi_1, \ldots, \phi_n$ inside the expectation, where randomness is based on the triplet set $\mathcal{S}$. However, the conditional expectation $\mathbb{E}_{\epsilon |\mathcal{S}}$ in (\ref{cond_exp}) is conditioned on $\mathcal{S}$. Note that the size of the Rademacher random vector $\epsilon$ is $|\mathcal{S}|$. We first focus on the conditional expectation:
\begin{eqnarray}
    \mathbb{E}_{\epsilon |\mathcal{S}} \left[\sup_L \sum_{t\in \mathcal{S}}\epsilon_t(\|L\phi_h-L\phi_i\|_\mathcal{H}^2-\|L\phi_h-L\phi_j\|_\mathcal{H}^2))\right]. \label{cond_exp2}
\end{eqnarray}
Consider the span of features $\phi_1, \ldots, \phi_n$ and call it $\mathcal{S}_\mathcal{X}$. Using Riesz's Representation Theorem, we can write $L\phi$ for any $\phi$ as follows:
\begin{eqnarray*}   L\phi=\sum_{k=1}^\infty\langle \phi, \tau_k\rangle_\mathcal{H} \mathbf{e}_k
\end{eqnarray*}
For the conditional expectation in (\ref{cond_exp2}), we can write each $\tau_k$ as the summation of $\tau_k'$ and $\tau_k^\perp$, where $\tau_k'$ represents the part lies in $\mathcal{S}_\mathcal{X}$ and $\tau_k^\perp$ is orthogonal to $\mathcal{S}_\mathcal{X}$. 
\begin{eqnarray*}
    \tau_k=\tau_k'+\tau_k^\perp.
\end{eqnarray*}
We can represent each $\tau_k'$ as $\sum_{j=1}^nv_{k,j}\psi_j$, where $\{\psi_1, \ldots, \psi_n\}$ is an orthonormal basis for the set $\{\phi_1, \ldots, \phi_n\}$ and $v_{k,j}\in \mathbb{R}, \forall k,j$. Therefore, for any $\phi_i, \phi_j \in \mathcal{S}_\mathcal{X}$,
\begin{eqnarray}
    \langle L\phi_i, L\phi_j\rangle_\mathcal{H} &=& \sum_{k=1}^\infty\langle \phi_i, \tau_k \rangle_\mathcal{H} \langle \phi_j, \tau_k \rangle_\mathcal{H} \nonumber
    \\ &=& \sum_{a=1}^n\sum_{b=1}^n \left(\sum_{k=1}^\infty v_{k,a}v_{k,b}\right)\langle \phi_i, \psi_a \rangle_\mathcal{H} \langle \phi_j, \psi_b \rangle_\mathcal{H} \nonumber \\
    &=& \varphi_i^T\mathbf{M}^{\mathcal{S}_\mathcal{X}} \varphi_j
    \label{Riesz_representation}
\end{eqnarray}
where $\varphi_{i}=[\langle \phi_{i}, \psi_1 \rangle, \langle \phi_{i}, \psi_2 \rangle, \ldots \langle \phi_{i}, \psi_n \rangle]^T$ and $\Mb^{\mathcal{S}_\mathcal{X}}_{i,j}=\sum_{k=1}^\infty v_{k,j}v_{k,i}$.  Note that $\Mb^{\mathcal{S}_\mathcal{X}}$ and $\{\varphi_1, \ldots, \varphi_n\}$ are functions of $\mathcal{S}$. Based on (\ref{Riesz_representation}), for $\phi_i,\phi_j \in \mathcal{S}$, we have
\begin{eqnarray*}
    &&\|L\phi_h-L\phi_i\|_\mathcal{H}^2-\|L\phi_h-L\phi_j\|_\mathcal{H}^2
    \\&=&(\varphi_{j}-\varphi_{i})^T\Mb^{\mathcal{S}_\mathcal{X}} (2\varphi_{h}-\varphi_{i}-\varphi_{j})
    \\ &=&\frac{1}{2}\left((\varphi_{j}-\varphi_{i})^T\Mb^{\mathcal{S}_\mathcal{X}} (2\varphi_{h}-\varphi_{i}-\varphi_{j})+(2\varphi_{h}-\varphi_{i}-\varphi_{j})^T\Mb^{\mathcal{S}_\mathcal{X}}(\varphi_{j}-\varphi_{i}) \right)
    \\&=&\frac{1}{2}\text{Tr}\left(\Mb^{\mathcal{S}_\mathcal{X}}(2\varphi_{h}-\varphi_{i}-\varphi_{j})(\varphi_{j}-\varphi_{i})^T+\Mb^{\mathcal{S}_\mathcal{X}}(\varphi_{j}-\varphi_{i})(2\varphi_{h}-\varphi_{i}-\varphi_{j})^T\right)
    \\&=&\text{Tr}\left(\Mb^{\mathcal{S}_\mathcal{X}}(\varphi_{h}\varphi_{j}^T+\varphi_{j}\varphi_{h}^T-\varphi_{h}\varphi_{i}^T-\varphi_{i}\varphi_{h}^T+\varphi_{i}\varphi_{i}^T-\varphi_{j}\varphi_{j}^T)\right)
\end{eqnarray*}
Suppose $\Kb_t=\varphi_{h}\varphi_{j}^T+\varphi_{j}\varphi_{h}^T-\varphi_{h}\varphi_{i}^T-\varphi_{i}\varphi_{h}^T+\varphi_{i}\varphi_{i}^T-\varphi_{j}\varphi_{j}^T$. Then, we have
\begin{eqnarray}
    \mathbb{E}_{\mathcal{S}}\left[\mathbb{E}_{\epsilon |\mathcal{S}} \left[\sup_L \sum_{t\in \mathcal{S}}\epsilon_t(\|L\phi_h-L\phi_i\|_\mathcal{H}^2-\|L\phi_h-L\phi_j\|_\mathcal{H}^2)\right] \right] \nonumber
    \\ =
    \mathbb{E}_{\mathcal{S}}\left[\mathbb{E}_{\epsilon |\mathcal{S}} \left[\sup_L \text{Tr}\left(\Mb^{\mathcal{S}_\mathcal{X}}\sum_{t\in \mathcal{S}}\epsilon_t \Kb_t\right)\right]\right]. \label{finite_kpca}
\end{eqnarray}
For the expression inside the expectations in (\ref{finite_kpca}), we have    
\begin{eqnarray}
\sup_L \text{Tr}\left(\Mb^{\mathcal{S}_\mathcal{X}}\sum_{t\in \mathcal{S}}\epsilon_t \Kb_t\right)
    &\overset{a}{\leq}& \sup_L\sum_{i=1}^r\sigma_i(\Mb^{\mathcal{S}_\mathcal{X}})\sigma_i\left(\sum_{t\in \mathcal{S}}\epsilon_t \Kb_t\right) \nonumber
    \\
    &\overset{b}{\leq}&  \sup_L \left[\|\Mb^{\mathcal{S}_\mathcal{X}}\|_{\text{F}} \|\sum_{t\in \mathcal{S}}\epsilon_t \Kb_t\|_{\text{F}}\right] \nonumber
     \\
    &\overset{c}{\leq}& \lambda_F\|\sum_{t\in \mathcal{S}}\epsilon_t \Kb_t\|_{\text{F}} \nonumber
    \\ 
    &{=}& \lambda_F\sqrt{\|\sum_{t\in \mathcal{S}}\epsilon_t\Kb_t\|_{\text{F}}^2} \label{bounding_trace}.
\end{eqnarray}
    Here, $(a)$ is from Von Neumann's trace inequality, $(b)$ is the result of Cauchy–Schwarz Inequality and we recall that $\|\Mb^{\mathcal{S}_\mathcal{X}}\|_{\text{F}}\leq\lambda_F$. Inserting (\ref{bounding_trace}) into (\ref{finite_kpca}), we can write
\begin{eqnarray*}
    \mathbb{E}_{\mathcal{S}}\left[\mathbb{E}_{\epsilon |\mathcal{S}} \left[\sup_L \sum_{t\in \mathcal{S}}\epsilon_t(\|L\phi_h-L\phi_i\|_\mathcal{H}^2-\|L\phi_h-L\phi_j\|_\mathcal{H}^2)\right] \right]
    \leq
\lambda_F\mathbb{E}_{\mathcal{S}}\left[\mathbb{E}_{\epsilon |\mathcal{S}} \left[\sqrt{\|\sum_{t\in \mathcal{S}}\epsilon_t\Kb_t\|_{\text{F}}^2}\right]\right].
\end{eqnarray*}
Then, we have
\begin{eqnarray}
\mathbb{E}_{\mathcal{S}}\left[\mathbb{E}_{\epsilon |\mathcal{S}} \left[\sqrt{\|\sum_{t\in \mathcal{S}}\epsilon_t\Kb_t\|_{\text{F}}^2}\right]\right]
       &\overset{a}{\leq}& \mathbb{E}_{\mathcal{S}}\left[\sqrt{\mathbb{E}_{\epsilon |\mathcal{S}}\left[\|\sum_{t\in \mathcal{S}}\epsilon_t\Kb_t\|_{\text{F}}^2\right]}\right]\nonumber
    \\
   &\overset{}{=}&\mathbb{E}_{\mathcal{S}}\left[ \sqrt{\mathbb{E}_{\epsilon |\mathcal{S}}\left[\langle\sum_{t\in \mathcal{S}}\epsilon_t\Kb_t, \sum_{t\in \mathcal{S}}\epsilon_t\Kb_t\rangle\right]}\right]\nonumber
    \\
   &\overset{}{=}& \mathbb{E}_{\mathcal{S}}\left[\sqrt{\mathbb{E}_{\epsilon |\mathcal{S}}\left[\sum_{t\in \mathcal{S}}\sum_{t'\in \mathcal{S}}\epsilon_t\epsilon_{t'}\langle\Kb_t, \Kb_{t'}\rangle\right]}\right]\nonumber
    \\
     &\overset{b}{=}& \mathbb{E}_{\mathcal{S}}\left[\sqrt{\mathbb{E}_{\epsilon |\mathcal{S}}\left[\sum_{t\in \mathcal{S}}\epsilon_t^2\langle\Kb_t, \Kb_t\rangle\right]}\right]\nonumber
    \\
    &\overset{}{=}& \mathbb{E}_{\mathcal{S}}\left[\sqrt{\sum_{t\in \mathcal{S}}\|\Kb_t\|_{\text{F}}^2}\right]\nonumber
    \\
    &\overset{}{\leq}& B^2\sqrt{|S|6} \label{expectation last}
\end{eqnarray}
where $(a)$ is from Jensen’s inequality where the expectation is over the randomness in $\epsilon_t$ and $(b)$ is due the fact that $\mathbb{E}(\epsilon_{t_1}\epsilon_{t_2})=0$ when $t_1\neq t_2$. For the last step, recall that $\Kb_t=\varphi_{h}\varphi_{j}^T+\varphi_{j}\varphi_{h}^T-\varphi_{h}\varphi_{i}^T-\varphi_{i}\varphi_{h}^T+\varphi_{i}\varphi_{i}^T-\varphi_{j}\varphi_{j}^T$. Then, we have
\begin{eqnarray}
    \|\Kb_t\|_{\text{F}}^2&\overset{a}{\leq}& 6\max_{i,j}\|\varphi_{i}\varphi_{j}^T\|_{\text{F}}^2 \nonumber
    \\ &\overset{b}{\leq}& 6B^4, \label{bounding_fro_norm of Kt}
\end{eqnarray}
where $(a)$ is by triangle inequality and $(b)$ follows from that fact that $\|\varphi_i\|_2 = \|\phi_i\|_\mathcal{H}\leq B$. Note that $\|\varphi_i\|_2 = \|\phi_i\|_\mathcal{H}$ is by definition, where $\varphi_i$ is defined via change of basis on the span $\mathcal{S}_\mathcal{X}$. Finally, from (\ref{Rademacher last}) and (\ref{expectation last}), we have
\begin{eqnarray*}
    R(\widehat{L}_0)-R(L^*)\leq 4\alpha B^2\lambda_F\sqrt{\frac{6}{|S|}}+2\ell\sqrt{\frac{2\gamma^2\ln{2/\delta}}{|\mathcal{S}|}},
\end{eqnarray*}
which completes the proof of Theorem \ref{thm:generalization_error_withbounded_Fro_norm}.

\begin{lemma}\label{lem:difference of distances}
    Let $\phi(\bx)$ be a feature map from $\mathbb{R}^d$ to $\mathcal{H}$ with $\|\phi(\bx)\|_\mathcal{H}\leq B$ for $\forall \bx$, and $L$ be a linear functional such that $L:\mathcal{H}\rightarrow \mathcal{H}$ and $\|L^\dagger L\|_{S_2} \leq \lambda_F
    $. Then, for any $\bx_h,\bx_i,\bx_j\in \mathbb{R}^d$, we have 
    \begin{eqnarray*}
        \|L\phi_h-L\phi_i\|_\mathcal{H}^2-\|L\phi_h-L\phi_j\|_\mathcal{H}^2\leq 6B^2 \lambda_F
    \end{eqnarray*}
\end{lemma}
\textbf{Proof of Lemma \ref{lem:difference of distances}}
First, note that 
\begin{eqnarray*}
    \langle L\phi_h, L\phi_j\rangle_\mathcal{H} &=& \langle \phi_h, L^\dagger L\phi_j\rangle_\mathcal{H}  
    \\&\overset{a}{\leq}& \|\phi_h\|_\mathcal{H} \|L^\dagger L\phi_j\|_\mathcal{H}
    \\&\overset{b}{\leq}& \|\phi_h\|_\mathcal{H} \|L^\dagger L\|_{S_\infty}\|\phi_j\|_\mathcal{H}
    \\&{\leq}& \|\phi_h\|_\mathcal{H} \|L^\dagger L\|_{S_2}\|\phi_j\|_\mathcal{H}
    \\&\leq& B^2\lambda_F,
\end{eqnarray*}
where $(a)$ is from Cauchy-Schwarz Inequality and $(b)$ is by definition of operator norm ($\|\cdot\|_{S_\infty}$). Then, we have 
\begin{eqnarray*}
    \|L\phi_h-L\phi_i\|_\mathcal{H}^2-\|L\phi_h-L\phi_j\|_\mathcal{H}^2&=&2\langle L\phi_h, L\phi_j\rangle_\mathcal{H}-2\langle L\phi_h, L\phi_i\rangle_\mathcal{H}+\langle L\phi_i, L\phi_i\rangle_\mathcal{H}-\langle L\phi_j, L\phi_j\rangle_\mathcal{H}
    \\ &\leq& 6B^2 \lambda_F.
\end{eqnarray*}

\subsection{Proof of Theorem \ref{thm:generalization_error_withbounded_Nuclear_norm}}


Recall that the only difference between the setting in Theorem \ref{thm:generalization_error_withbounded_Fro_norm} and the setting in Theorem \ref{thm:generalization_error_withbounded_Nuclear_norm} is the constraint set. We replace the constraints $\|\mathcal{P}^\dagger_{\mathcal{S}_\mathcal{X}}L^\dagger L\mathcal{P}_{\mathcal{S}_\mathcal{X}}\|_{S_2}\leq \lambda_F$ and $\|\Mb\|_F\leq \lambda_F$
with $\|\mathcal{P}^\dagger_{\mathcal{S}_\mathcal{X}}L^\dagger L\mathcal{P}_{\mathcal{S}_\mathcal{X}}\|_{S_1}\leq \lambda_*$ and $\|\Mb\|_*\leq \lambda_*$ respectively. We update definitions accordingly. Then, the proof follows the same steps  with the proof of Theorem \ref{thm:generalization_error_withbounded_Fro_norm} until (\ref{finite_kpca}), where we have 
\begin{eqnarray}
R(\widehat{L}_{n_0})-R(L^*_n)
    &\leq& \frac{4\alpha}{|\mathcal{S}|} \mathbb{E}_{\mathcal{S}}\left[\mathbb{E}_{\epsilon |\mathcal{S}} \left[\sup_L \text{Tr}\left(\Mb^{\mathcal{S}_\mathcal{X}}\sum_{t\in \mathcal{S}}\epsilon_t \Kb_t\right)\right]\right] + \beta\sqrt{\frac{2\ln{2/\delta}}{|\mathcal{S}|}}
   \label{kpca bound_nuclear norm}
\end{eqnarray}
We focus on the expression inside the expectations and we can write    
\begin{eqnarray}
\sup_L \text{Tr}\left(\Mb^{\mathcal{S}_\mathcal{X}}\sum_{t\in \mathcal{S}}\epsilon_t \Kb_t\right)
    &\overset{a}{\leq}& \sup_L\|\Mb^{\mathcal{S}_\mathcal{X}}\|\left\|\sum_{t\in \mathcal{S}}\epsilon_t \Kb_t\right\| \nonumber
    \\
    &{\leq}&  \sup_L\|\Mb^{\mathcal{S}_\mathcal{X}}\|_*\left\|\sum_{t\in \mathcal{S}}\epsilon_t \Kb_t\right\| \nonumber
     \\
    &\overset{b}{\leq}& \lambda_*\|\sum_{t\in \mathcal{S}}\epsilon_t \Kb_t\|  \label{bounding_trace_nuclear norm}
\end{eqnarray}
  Here, $(a)$ is from Hölder's Ineqaulity for Schatten norms and we recall that $\|\Mb^{\mathcal{S}_\mathcal{X}}\|_{*}\leq\lambda_*$ for $(b)$. Inserting (\ref{bounding_trace_nuclear norm}) into the expectations in (\ref{kpca bound_nuclear norm}), we can write
\begin{eqnarray*}
    \mathbb{E}_{\mathcal{S}}\left[\mathbb{E}_{\epsilon |\mathcal{S}} \left[\sup_L \sum_{t\in \mathcal{S}}\epsilon_t(\|L\phi_h-L\phi_i\|_\mathcal{H}^2-\|L\phi_h-L\phi_j\|_\mathcal{H}^2)\right] \right]
    \leq
\lambda_*\mathbb{E}_{\mathcal{S}}\left[\mathbb{E}_{\epsilon |\mathcal{S}} \left[\left\|\sum_{t\in \mathcal{S}}\epsilon_t \Kb_t\right\|\right]\right].
\end{eqnarray*}
Then, we have
\begin{eqnarray}
\lambda_*\mathbb{E}_{\mathcal{S}}\left[\mathbb{E}_{\epsilon |\mathcal{S}}\left[\left\|\sum_{t\in \mathcal{S}}\epsilon_t \Kb_t\right\|\right]\right]
   &\overset{c}{\leq}&\lambda_*\mathbb{E}_{\mathcal{S}}\left[\sqrt{2\left\|\sum_{t=1}^{|\mathcal{S}|}\mathbb{E}_{\mathcal{\epsilon}}\left[\Kb_t^2\right]\right\|\log 3|\mathcal{S}|}+2\log 3|\mathcal{S}|\right] \nonumber
    \\
     &\overset{d}{\leq}& \lambda_*\mathbb{E}_{\mathcal{S}}\left[\sqrt{12B^4|\mathcal{S}|\log 3|\mathcal{S}| } +2\log 3|\mathcal{S}|\right]\nonumber
         \\
     &{=}& \lambda_*\left(\sqrt{12B^4|\mathcal{S}|\log 3|\mathcal{S}| } +2\log 3|\mathcal{S}|\right) \label{matrixBern}
\end{eqnarray}
where we apply a matrix Bernstein bound to get $(c)$ (see Theorem 6.6.1 in \cite{tropp2015introduction}) and $(d)$ follows from (\ref{bounding_fro_norm of Kt}). Lastly, from (\ref{kpca bound_nuclear norm}) and (\ref{matrixBern}), we have 
\begin{eqnarray*}
    R(\widehat{L}_{n_0})-R(L^*_n)\leq 4\alpha\lambda_*\left(B^2\sqrt{12\frac{\log 3|\mathcal{S}|}{|\mathcal{S}|} } +\frac{2\log 3|\mathcal{S}|}{|\mathcal{S}|}\right)+12\alpha B^2\lambda_*\sqrt{\frac{2\ln{2/\delta}}{|\mathcal{S}|}},
\end{eqnarray*}
which completes the proof of Theorem \ref{thm:generalization_error_withbounded_Nuclear_norm}.
\subsection{Proof of Lemma \ref{lem:relating norms}}
\label{sec:proof of lem relating norms}
Let $\psi_1, \ldots, \psi_n \in \mathcal{H}$ denote the KPCA directions which span $\mathcal{S}_\mathcal{X}$ such that $\langle\psi_i, \phi_j\rangle=(\varphi_j)_i\in \mathbb{R}$, where $(v)_i$ denotes the $i^{th}$ entry of vector $v$. Furthermore, let $\mathbf{B}_i$, denote the $ith$ row of a matrix $\mathbf{B}$. For any $L:\mathcal{H} \rightarrow \mathcal{H}$ and $\phi_x \in \mathcal{H}$ we may write $L \mathcal{P}_{\mathcal{S}_\mathcal{X}}\phi_x = \sum_{i=1}^n\sum_{j=1}^nw_{i,j}\Psi_i\otimes \Psi_j \phi_x$, where $\Psi_i\otimes \Psi_j \phi_x = \langle \Psi_j, \phi_x\rangle_\mathcal{H}\Psi_i$. Let $\Wb$ be the matrix of $w_{ij}$ weights. Lastly, let $a^Tb$ denote the standard Euclidean inner product for $a, b \in \mathbb{R}^n$. Then, for $\phi_x, \phi_y \in \mathcal{H}$,
\begin{eqnarray}
    \|L\mathcal{P}_{\mathcal{S}_\mathcal{X}}\phi_x-L\mathcal{P}_{\mathcal{S}_\mathcal{X}}\phi_y\|_\mathcal{H}^2 
    &=&\langle L\mathcal{P}_{\mathcal{S}_\mathcal{X}}\phi_x-L\mathcal{P}_{\mathcal{S}_\mathcal{X}}\phi_y, L\mathcal{P}_{\mathcal{S}_\mathcal{X}}\phi_x-L\mathcal{P}_{\mathcal{S}_\mathcal{X}}\phi_y\rangle \label{firstline Lem1}
    \\ &=& \left\langle\sum_{i=1}^n\sum_{j=1}^nw_{i,j}\Psi_i\otimes \Psi_j (\phi_x-\phi_y), \sum_{i=1}^n\sum_{j=1}^nw_{i,j}\Psi_i\otimes \Psi_j (\phi_x-\phi_y)\right\rangle_\mathcal{H} \nonumber
    \\ &=&\left\langle\sum_{i=1}^n\sum_{j=1}^nw_{i,j}\langle\Psi_j, \phi_x-\phi_y\rangle_\mathcal{H}\Psi_i, \sum_{i=1}^n\sum_{j=1}^nw_{i,j}\langle\Psi_j, \phi_x-\phi_y\rangle_\mathcal{H}\Psi_i\right \rangle_\mathcal{H} \nonumber
    \\ &=& \left\langle\sum_{i=1}^n\sum_{j=1}^nw_{i,j}\left((\varphi_\bx)_j-(\varphi_\by)_j\right)\Psi_i, \sum_{i=1}^n\sum_{j=1}^nw_{i,j}\left((\varphi_\bx)_j-(\varphi_\by)_j\right)\Psi_i \right\rangle_\mathcal{H} \nonumber
    \\ &=& \left\langle \sum_{i=1}^n \Wb^T_i(\varphi_\bx-\varphi_\by)\Psi_i, \sum_{i=1}^n \Wb^T_i(\varphi_\bx-\varphi_\by)\Psi_i \right \rangle_\mathcal{H} \nonumber
    \\ &=& \sum_{i=1}^n \Wb^T_i(\varphi_\bx-\varphi_\by) \left\langle \Psi_i, \sum_{j=1}^n\Wb^T_j(\varphi_\bc-\varphi_\by)\Psi_j\right\rangle_\mathcal{H} \nonumber
    \\ &=& \sum_{i=1}^n \left(\Wb^T_i(\varphi_\bx-\varphi_\by)\right)^2\langle\Psi_i,\Psi_i\rangle_\mathcal{H} \nonumber
    \\ &=& \sum_{i=1}^n (\varphi_\bx-\varphi_\by)^T\Wb_i\Wb_i^T(\varphi_\bx-\varphi_\by) \nonumber
    \\ &=& (\varphi_\bx-\varphi_\by)^T\Wb\Wb^T(\varphi_\bx-\varphi_\by) \nonumber
    \\ &=& \|\varphi_\bx-\varphi_\by\|_\Mb^2 \label{lastline lem1}
\end{eqnarray}
where in the final step we have defined $\Mb := \Wb \Wb^T$. Then the eigenvalues of $\Mb$ are equal to the square of the singular values of $\Wb$. In general we note that the eigenvalues of $(L\mathcal{P}_{\mathcal{S}_\mathcal{X}})^\dagger L\mathcal{P}_{\mathcal{S}_\mathcal{X}}$ are equal to the eigenvalues of $\Mb$ where $L^\dagger$ denote the adjoint of $L$. Note that $\|L\phi_x-L\phi_y\|_\mathcal{H}^2=\|L\mathcal{P}_{\mathcal{S}_\mathcal{X}}\phi_x-L\mathcal{P}_{\mathcal{S}_\mathcal{X}}\phi_y\|_\mathcal{H}^2$ for $\phi_x,\phi_y \in \mathcal{S}_\mathcal{X}$. Hence, we have $\|L\phi_\xb\|_\mathcal{H}=\|\varphi_\xb\|_\Mb$ for any $\phi_x\in \mathcal{S}_\mathcal{X}$ from (\ref{lastline lem1}).
\subsection{Proof of Proposition \ref{prop:operationalizing}}
From Lemma \ref{lem:relating norms}, we have $\|\mathcal{P}^\dagger_{\mathcal{S}_\mathcal{X}}L^\dagger L\mathcal{P}_{\mathcal{S}_\mathcal{X}}\|_{S_p}=\|\Mb\|_p$ $\forall p$. Similarly, from the fact that $\|L\phi_\xb\|_\mathcal{H}=\|\varphi_\xb\|_\Mb$ (see Lemma \ref{lem:relating norms}), we have $|\|L\phi_h-L\phi_i\|^2-\|L\phi_h-L\phi_j\|^2|=|\|\varphi_h-\varphi_i\|_\Mb^2-\|\varphi_h-\varphi_j\|_\Mb^2|$ within $\mathcal{S}_{\mathcal{X}}$. Then, from Proposition \ref{prop:representer theorem}, we conclude that (\ref{opt-P3}) and $(\ref{opt-P4})$ have the same optimal value. Therefore, we have that
\begin{equation}
\begin{aligned}
\min_{L} \quad & \widehat{R}_\mathcal{S}(L)\\
\textrm{s.t.} \quad & \|\mathcal{P}_{\mathcal{S}_\mathcal{X}}^\dagger L^\dagger L\mathcal{P}_{\mathcal{S}_\mathcal{X}}\|_{S_2}\leq \lambda_F    \\
\end{aligned} \tag{P3}
\end{equation}
is equal to
\begin{equation}
\begin{aligned}
\min_{\Mb} \quad & \widehat{\overline{R}}_\mathcal{S}(\Mb)\\
\textrm{s.t.} \quad &   \|\Mb\|_F \leq \lambda_F    \\
  &  \Mb \succeq 0,    \\
\end{aligned}\tag{P4}
\end{equation}
By definition, $\widehat{\Lb}_0$ is an optimal solution for (\ref{opt-P3}) and $\widehat{\Mb}$ is the optimal solution for (\ref{opt-P4}). Recall that there exists a psd matrix $\Mb$ for each pair of $L$ and $\mathcal{S}_\mathcal{X}$ from Lemma \ref{lem:relating norms}.

For the construction of $\widehat{L}_0$ from $\widehat{\Mb}$, we follow similar lines with the proof of Lemma \ref{lem:relating norms}.  $\widehat{L}_0$ is defined reversing equalities in Section \ref{sec:proof of lem relating norms} for $\Mb = \widehat{\Mb}$, from (\ref{firstline Lem1}) to (\ref{lastline lem1}). Therefore, we observe that $\widehat{\Mb}$ is the corresponding psd matrix for the pair $(\widehat{L}_0, \mathcal{S}_\mathcal{X})$. This is actually true for any $L$ provided that $L\mathcal{P}_{\mathcal{S}_\mathcal{X}}=\widehat{L}_0$.

\section{Discussion} \label{appendix:discussion}
Our results extend the linear metric setting of Mason et al. \cite{mason2017learning} in two key ways: First, our main results provide generalization error and sample complexity bounds for the kernelized metric learning from triplet comparisons. Second, the linear metric learning analysis of Mason et al. \cite{mason2017learning} requires that the number of items, $n$, be larger than the dimensionality, $d$, which limits its applicability. In contrast, our analysis, which also considers linear kernels, offers a more general framework, even for linear metric learning from triplet comparisons.

Mason et al. \cite{mason2017learning} consider a fixed set of items in $\mathbb{R}^d$
and derive generalization bounds based on selecting triplets uniformly from those that can be generated from the fixed item set. Their analysis exploits the fact that the item set is fixed and requires that the number of items n is larger than the dimensionality d, which limits its applicability. Also note that, the true risk of Mason et al. \cite{mason2017learning} is defined with respect to a discrete uniform distribution over $n{\binom{n-1}{2}}$ triplets possible from the fixed set of $n$ items.

Our setting differs significantly from Mason et al. \cite{mason2017learning} in the following aspects: We do not assume a fixed set of items, which would otherwise restrict generalization only to the triplets drawn from this fixed set. Instead, in our setting, each triplet query involves items drawn iid from an unknown distribution $\mathcal{D}$. Our true risk is defined over this unknown distribution and the generalization bounds hold for triplets chosen from this distribution. Thus, our analysis also extends the generalization results even for the linear kernel case in high dimensions (large $d$) apart from generalizing to infinite-dimensional RKHS.

Given the difference in settings, the proof technique we use differs from Mason et al. \cite{mason2017learning}. To derive our sample complexity results, we turn our attention to the metric and exploit the fact that the true metric ${L}^*$ has a bounded Schatten $p$-norm, which constrains how $L$ interacts with any random data. We use this constraint in conjunction with the Riesz Representation Theorem to further refine our analysis.

\end{document}